\documentclass[11pt,a4paper]{article}

\usepackage{xpeng-template}

\usepackage{tikz}
\usepackage{pgfplots}
\usetikzlibrary{plotmarks, calc, positioning, fit, arrows.meta, backgrounds}
\pgfplotsset{compat=1.18}
\usepackage{placeins}
\usepackage{hyperref}


\definecolor{EdgeBg}{HTML}{EDF7ED}
\definecolor{MidBg}{HTML}{E8F0FE}
\definecolor{MassiveBg}{HTML}{FFF3E0}
\definecolor{ClosedBg}{HTML}{F3E8FD}
\definecolor{FRtint}{HTML}{FDE8E8}
\definecolor{NRtint}{HTML}{E6F4F1}
\definecolor{PolicyPoint}{HTML}{0077BB}
\definecolor{SelectedPolicy}{HTML}{CC3311}

\definecolor{LocS2}{HTML}{4E46C9}      
\definecolor{LocS1}{HTML}{0E7A6E}      
\definecolor{LocS0}{HTML}{B25E12}      
\definecolor{LocHandoff}{HTML}{5E6A82} 
\definecolor{LocReport}{HTML}{9AA6B6}  
\definecolor{LocInk}{HTML}{1B2230}
\definecolor{LocInkSoft}{HTML}{5A6577}

\definecolor{CleS2bg}{HTML}{EFEFFB}
\definecolor{CleS2bd}{HTML}{C9C7F0}
\definecolor{CleS1bg}{HTML}{E5F3F0}
\definecolor{CleS0bg}{HTML}{FBF1E3}
\definecolor{CleUR}{HTML}{475569}
\definecolor{CleURbg}{HTML}{EEF1F6}
\definecolor{CleThread}{HTML}{5E6A82}
\definecolor{CleHair}{HTML}{EBEEF6}
\pgfdeclarelayer{cle-bg}
\pgfdeclarelayer{cle-wires}
\newcommand{\clesw}[2]{\tikz[baseline=-0.45ex]{\draw[#1,#2,fill=#1!30,rounded corners=1.2pt,%
  line width=1pt] (0,0) rectangle (0.62em,0.42em);}}

\newcommand{\pmark}{\tikz[baseline=-0.55ex,scale=0.07]{\fill (0,0) -- (90:1) arc[start angle=90,end angle=270,radius=1] -- cycle; \draw (0,0) circle (1);}}

\xpengaffil{XPENG Robotics}
\xpengcorrespondence{Jie Chen}{chenj81@xiaopeng.com}

\title{DeepInsight: A Unified Evaluation Infrastructure Across the Physical AI Stack}
\author{Siyi Li, Chunyu Sun, Jiahao Zhang, Yuchen Kang, Wuliang Wang, Yu Qiu,
Rui Jiang, Haitao Cui, Jie Chen\textsuperscript{\textdagger}}
\date{}

\begin{document}

%

\begin{xpenghero}
\begin{abstract}
Evaluating a Physical AI stack spans operators that differ by more than three orders of magnitude---from a single foundation-model decoding step to thousands of physics ticks of whole-body control---varying orthogonally in modality, reward semantics, and resource profile. No existing framework spans this range, so the stack is evaluated today by stitching together separate harnesses that share neither runtime nor scoring, preserving each segment's local validity but losing the shared identity needed to diagnose cross-layer regressions. We present DeepInsight, an evaluation infrastructure that serves this full spectrum on a single runtime. Rather than homogenize the regimes, it preserves their heterogeneity behind three narrow abstractions---task, resource, and result---each realized as one invariant shared by every subsystem: one episode driver, one resource-handle protocol implemented by every expensive backend (LLM inference and sandboxed runtimes alike), and one trace identity scheme under which every event is written.
Deployed in production across all three layers of an embodied humanoid stack, this single set of invariants onboards new benchmarks largely by configuration. Where mature peer orchestrators exist---at the foundation-model end---it reproduces published references and peer-framework readings within their own spread, runs the same suites faster on a single node, and scales near-linearly across nodes. Its distinctive return is diagnostic: because every layer writes into one shared trace, a regression that begins in one layer and surfaces in another stays localizable on that trace---a cross-layer payoff no federation of per-segment harnesses can reproduce.
\end{abstract}
\end{xpenghero}

\section{Introduction}
\label{sec:intro}

Evaluation for a Physical AI stack is heterogeneous at the level of the operators it must drive. At one end of the workload sit foundation models: short episodes---often a single decoding step, sometimes tens of tool-calling turns---driven by throughput-bound inference and scored by exact-match or model-based judgment. At the other end sit whole-body control policies: episodes spanning hundreds to thousands of physics ticks, driven by physics-bound simulation, and scored by trajectory-analytic conditions on balance, contact, and tracking. Between these endpoints lies an inventory of embodied evaluation needs---manipulation policies, navigation stacks and so on. This paper introduces \textbf{DeepInsight}, an evaluation infrastructure that serves this full spectrum on a single runtime.

\begin{figure}[t]
  \centering
  \includegraphics[width=0.95\linewidth]{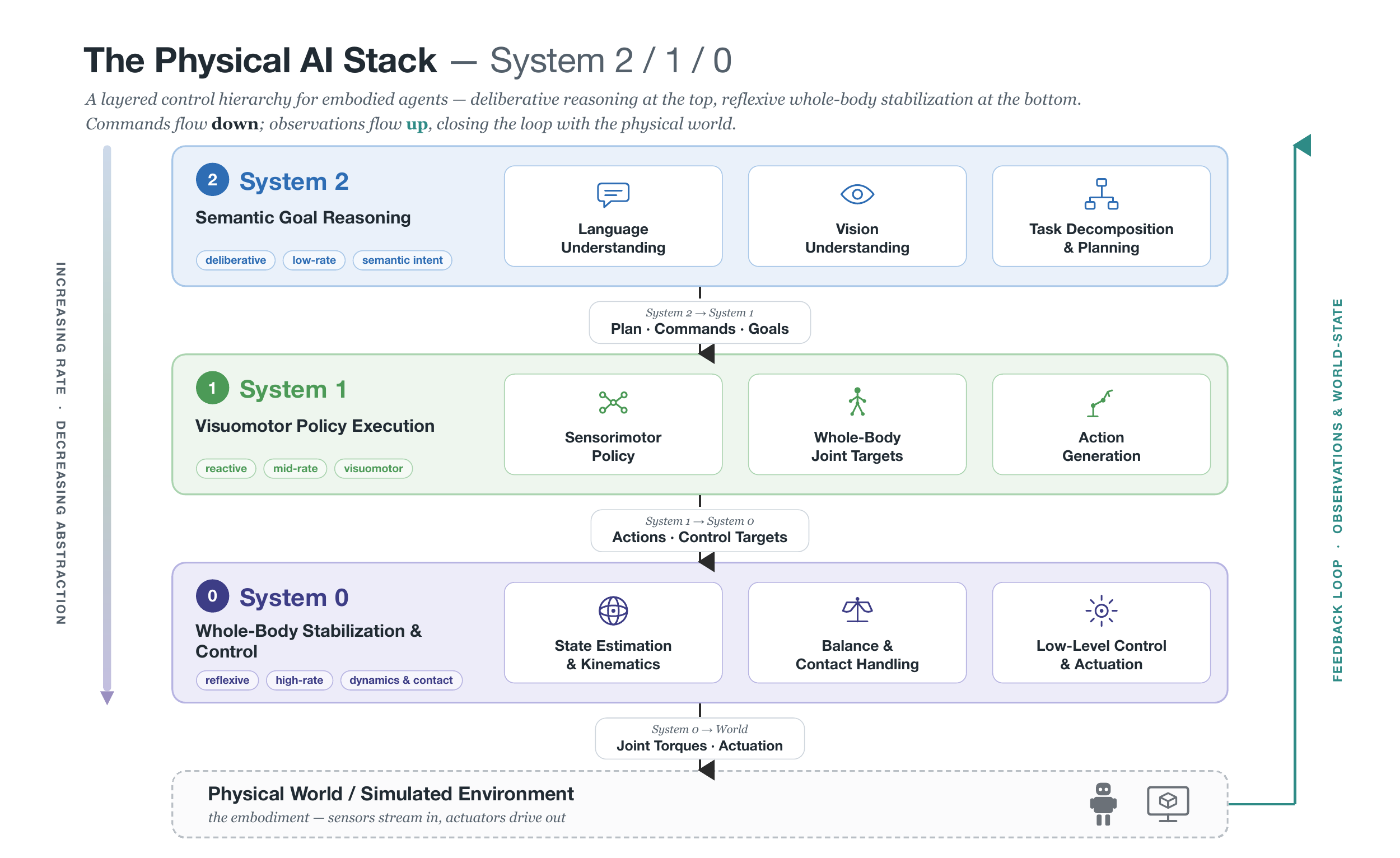}
  \caption{Physical AI stack considered by DeepInsight.}
  \label{fig:physical-ai-stack}
\end{figure}

DeepInsight is not a universal definition of Physical AI; it is an evaluation substrate for a specific embodied humanoid stack. Following recent industrial humanoids~\citep{figureai2026helix02}, we take this stack to be three layers: semantic goal reasoning at \emph{System~2} (foundation-model evaluation), visuomotor policy execution at \emph{System~1} (navigation/manipulation evaluation), and whole-body stabilization and control at \emph{System~0} (whole-body-control evaluation). Figure~\ref{fig:physical-ai-stack} sketches the stack; DeepInsight serves evaluation across all three layers in production. We use ``Physical AI'' throughout this paper as a label for this evaluation \emph{spectrum}, in the sense of the operator continuum just described.

The heterogeneity of this spectrum is not incidental; it is the engineering substance of the problem. Episode lengths span more than three orders of magnitude, from a single decoding step to thousands of physics ticks. Observation modalities range over text, image, audio, and continuous physics state. Reward semantics range from exact string match through model-based judgment to trajectory-analytic termination. Resource profiles range from GPU-bound model inference, through I/O- and CPU-bound sandboxed execution, to compute-bound parallel simulation that may be CPU- or GPU-resident. These axes do not collapse---no single representative point captures the spectrum---and they vary essentially orthogonally: a foundation-model agentic task may share its episode length with a manipulation policy while sharing nothing of its resource profile. Building a single evaluation infrastructure on this spectrum means an architecture that absorbs heterogeneity across the full range without forcing one segment's assumptions on another, and that lets new tasks anywhere along the spectrum enter through extension rather than reimplementation.


Heterogeneity does not by itself demand a unified evaluation infrastructure; one could in principle build separate harnesses for each segment of the spectrum and let them evolve independently. The case for unification is that, in a deployed Physical AI stack, the failures of these layers are coupled. A semantic planner's error changes the distribution seen by the visuomotor policy; a policy's hesitation changes the stabilizer's operating regime; a stabilizer's recovery behavior changes what the higher layers can attempt. Evaluating these layers in separate harnesses preserves their local benchmark validity but destroys the shared run identity, resource accounting, and trace continuity needed to diagnose cross-layer regressions. DeepInsight therefore unifies evaluation not by pretending the regimes are homogeneous---they plainly are not---but by preserving their heterogeneity behind common task, resource, and result interfaces.

DeepInsight's architecture is organized around three abstractions, each chosen to absorb a different class of spectrum heterogeneity. The \textbf{task abstraction} absorbs heterogeneity in episode shape, observation modality, reward semantics, and termination, expressing tasks from across the spectrum on the same runtime through a narrow \texttt{reset}/\texttt{step} interface and a per-episode handle on which all transient state lives. The \textbf{resource abstraction} absorbs heterogeneity in backend resource profile and operational irregularity, detaching the expensive resource classes that drive evaluation cost---language-model inference and sandboxed runtimes (covering code containers and physics simulators alike)---from the orchestrator, so that the operational unruliness of any backend cannot consume the orchestrator's async budget. The \textbf{result abstraction} absorbs heterogeneity in event type across the layers, recording the runtime's dialogue, the judge's rationale, the resource layer's lease and inference events, and the simulator's trajectory under one schema and one identity scheme, so that an aggregate score remains a queryable join over its constituents. Each abstraction carries a primary load: the task abstraction underwrites span coverage by letting heterogeneous episodes share one runtime, the resource abstraction underwrites in-regime throughput by detaching expensive resources from the orchestrator, and the result abstraction underwrites compositional extensibility by admitting new event types and analyses through the same trace. Concretely, the claim this paper defends is that a heterogeneous Physical AI evaluation workload can be carried by one episode driver, one resource-handle protocol, and one trace identity scheme without sacrificing benchmark fidelity, in-regime throughput, or the ability to localize cross-layer regressions on the trace.

\paragraph{Contributions.}
DeepInsight absorbs the heterogeneity of the Physical AI spectrum behind three abstractions---task, resource, and result---each realized as one invariant every subsystem shares: one episode driver, one resource-handle protocol, and one trace identity scheme. Where existing frameworks each cover a single segment, these three invariants carry the spectrum end to end---foundation-model decoding through whole-body control---on a single runtime.

We defend this in two parts, mirroring the two halves of the evaluation:
\begin{itemize}\setlength\itemsep{0.15em}
  \item \textbf{Production-grade where peers exist.} At the foundation-model end---the only segment with mature peer orchestrators---DeepInsight reproduces published references and peer-framework readings within a stated error budget, runs the same suites faster than the strongest single-regime baseline on a single node, and scales near-linearly across nodes; these gains follow from architecture-level mechanisms that are not segment-specific (Section~\ref{sec:evaluation}).
  \item \textbf{Full-stack reach, and cross-layer diagnosis.} The same runtime carries the rest of the stack, where no peer orchestrator exists: a progression of case studies reaches closed-loop simulation and trajectory-analytic release evaluation, and culminates in a composed System~2--1--0 task where a regression surfacing in one layer is diagnosed at its origin in another---a cross-layer localization on one shared trace that no federation of per-segment harnesses can reproduce (Section~\ref{sec:case-studies}).
\end{itemize}
\section{Related Work}
\label{sec:related-work}

\paragraph{Benchmarks across the Physical AI stack.}
Each layer of the embodied humanoid stack has a mature, internally coherent benchmark ecosystem, developed by largely disjoint communities. Projected onto the operational spectrum sketched in Section~\ref{sec:intro}, these benchmarks occupy distinct segments rather than a common ground. At System~2, static knowledge-and-reasoning QA---MMLU~\citep{hendrycks2021mmlu}, GSM8K~\citep{cobbe2021gsm8k}, HumanEval~\citep{chen2021humaneval}---sits at the spectrum's short-episode, exact-match, throughput-bound end: one decoding step per sample, deterministic scoring, and an LLM-inference resource profile. Long-horizon agentic harnesses---SWE-bench~\citep{jimenez2024swebench}, GAIA~\citep{mialon2024gaia}, OSWorld~\citep{xie2024osworld}, $\tau$-bench~\citep{yao2024taubench}, WebArena~\citep{zhou2024webarena}---move one step inward: episodes of tens to hundreds of turns, mixed exact-match and model-based judgment, and a resource profile that adds sandboxed runtimes to LLM inference. At System~1, visuomotor policy benchmarks---CALVIN~\citep{mees2022calvin}, LIBERO~\citep{liu2023libero}, Meta-World~\citep{yu2020metaworld}, RLBench~\citep{james2020rlbench}, Open X-Embodiment~\citep{oxe2024}, SimplerEnv~\citep{li2024simplerenv}---occupy a middle band: episodes of tens to a few hundred control steps, scoring that blends task-completion booleans with trajectory features, and a resource profile dominated by simulation rather than inference. At System~0, whole-body stabilization and locomotion---HumanoidBench~\citep{sferrazza2024humanoidbench}, RoboHive~\citep{kumar2023robohive}, Isaac Lab~\citep{mittal2023isaaclab}---anchor the long-episode, trajectory-analytic, physics-bound end: hundreds to thousands of physics ticks per rollout, continuous reward, and parallel simulation as the dominant cost. Each benchmark is internally consistent within its segment; signals across segments are not directly comparable because the operators producing them are not the same operators. The structural observation that motivates DeepInsight is simpler still: \emph{no single benchmark spans the spectrum from end to end}, and the spaces between adjacent segments are spaces where evaluation needs accumulate without a common substrate.

\paragraph{Frameworks for evaluation orchestration.}
A parallel infrastructure literature has emerged around how to run these benchmarks at scale, and each framework in it bakes in assumptions whose validity is local to its segment. The lm-evaluation-harness~\citep{sutawika2026lmevaluationharness} assumes short episodes, static datasets, and deterministic scorers; this makes it efficient on the short-episode, exact-match segment and incompatible with everything else. OpenCompass~\citep{opencompass2023} assumes per-sample cost is predictable and stages are homogeneous, which lets it shard statically across Slurm jobs but disqualifies it for workloads where stage cost is dynamic or stages are heterogeneous. HELM~\citep{liang2022helm} organizes 42 scenarios under a Scenario~$\times$~Metric~$\times$~Adapter abstraction that is methodological rather than operational; its operational footprint remains within text and short-horizon multimodal QA. VLMEvalKit~\citep{duan2024vlmevalkit} and lmms-eval~\citep{zhang2024lmmseval} extend coverage to 80--100+ vision--language benchmarks each, but inherit the single-shot generation assumption of their text antecedents and stay anchored at the short-episode end. Inspect AI~\citep{ukaisi2024inspectai} relaxes that assumption: its \texttt{Task = Dataset + Solver + Scorer} abstraction with asynchronous solver--scorer execution makes it the strongest open-source baseline for our short- and mid-episode comparisons in Section~\ref{sec:exp-efficiency}, but the \texttt{Task} contract is still scoped to foundation-model evaluation, with no first-class notion of long physics-tick episodes or trajectory-analytic scoring. From the opposite end, Isaac Lab~\citep{mittal2023isaaclab} provides robust parallel-simulation infrastructure for whole-body control evaluation, but it is a simulator framework, not a model/sandbox/judge orchestration framework: bringing language-model agents or sandbox-coupled tasks under its execution model would mean crossing its abstraction boundary, not extending it.

\begin{table}[t]
  \centering
  \caption{Coverage of the Physical AI evaluation spectrum across orchestration frameworks. Rows are spectrum axes; columns are frameworks. Check: first-class support; half-filled circle: partial coverage or support via extension; cross: not supported or out of scope. DeepInsight provides first-class support across the spectrum's full range; the contrast with single-segment frameworks is the structural observation that motivates the rest of the paper.}
  \label{tab:framework-comparison}
  \scriptsize
  \setlength{\tabcolsep}{3pt}
  \renewcommand{\arraystretch}{1.10}
  \resizebox{\linewidth}{!}{%
    \begin{tabular}{lccccccc}
      \toprule
      Spectrum axis & lm-eval & OpenCompass & HELM & VLMEvalKit & lmms-eval & Inspect AI & \textbf{DeepInsight} \\
      \midrule
      \multicolumn{8}{l}{\textbf{\emph{Episode length}}} \\
      $\le 1$ step (single decode) & \cmark & \cmark & \cmark & \cmark & \cmark & \cmark & \cmark \\
      2--50 steps (multi-turn, tool use) & \xmark & \pmark & \xmark & \xmark & \xmark & \cmark & \cmark \\
      50--500 steps (sandboxed agent, manipulation) & \xmark & \xmark & \xmark & \xmark & \xmark & \pmark & \cmark \\
      $\ge 500$ physics ticks (whole-body, locomotion) & \xmark & \xmark & \xmark & \xmark & \xmark & \xmark & \cmark \\
      \midrule
      \multicolumn{8}{l}{\textbf{\emph{Reward semantics}}} \\
      Exact / log-likelihood / rule-based & \cmark & \cmark & \cmark & \cmark & \cmark & \cmark & \cmark \\
      Model-based judge & \xmark & \pmark & \pmark & \pmark & \pmark & \cmark & \cmark \\
      Trajectory-analytic (continuous, in-env) & \xmark & \xmark & \xmark & \xmark & \xmark & \xmark & \cmark \\
      \midrule
      \multicolumn{8}{l}{\textbf{\emph{Backend resource profile}}} \\
      LLM inference only & \cmark & \cmark & \cmark & \cmark & \cmark & \cmark & \cmark \\
      + Sandboxed runtimes & \xmark & \xmark & \xmark & \xmark & \xmark & \cmark & \cmark \\
      + Physics-bound parallel simulation & \xmark & \xmark & \xmark & \xmark & \xmark & \xmark & \cmark \\
      \midrule
      \multicolumn{8}{l}{\textbf{\emph{Input modality}}} \\
      Text & \cmark & \cmark & \cmark & \cmark & \cmark & \cmark & \cmark \\
      + Vision & \xmark & \pmark & \pmark & \cmark & \cmark & \pmark & \cmark \\
      + Audio & \xmark & \xmark & \pmark & \xmark & \cmark & \xmark & \cmark \\
      + Physics state & \xmark & \xmark & \xmark & \xmark & \xmark & \xmark & \cmark \\
      \midrule
      \multicolumn{8}{l}{\textbf{\emph{Execution model}}} \\
      Multi-node execution & \xmark & \cmark & \xmark & \xmark & \xmark & \pmark & \cmark \\
      Stage-decoupled asynchronous & \xmark & \xmark & \xmark & \xmark & \xmark & \pmark & \cmark \\
      \bottomrule
    \end{tabular}%
  }
\end{table}

\paragraph{The shape of the gap.}
Table~\ref{tab:framework-comparison} is not a list of inadequacies: each framework is a well-engineered substrate within its regime. What it shows is that the abstractions these frameworks adopt are themselves regime-local---each shaped by the operator assumptions of its segment, each disqualified outside it. Extending any one of them across regime boundaries would mean discarding the assumptions that gave it efficiency in the first place, leaving extension and reimplementation indistinguishable. DeepInsight is designed against this structural gap, not against any particular benchmark; its architecture, developed in Section~\ref{sec:system-design}, follows from the shape of the gap.

\section{System Design}
\label{sec:system-design}

This section presents DeepInsight's architecture. Section~3.1 gives the system in overview; Section~3.2, Section~3.3, and Section~3.4 each take up one of its three abstractions and develop the engineering choices that abstraction forces.

\subsection{Architecture Overview}
\label{sec:architecture-overview}

DeepInsight's architecture rests on a single premise: the operator heterogeneity that defines the Physical AI evaluation spectrum must be absorbed by \emph{abstractions}---narrow interfaces across which one axis of heterogeneity is hidden from all the others. Three such abstractions organize the system; the remainder of this section argues that no smaller set suffices.

The first is the \textbf{task abstraction}: the interface between \emph{what a task is} and \emph{how the runtime drives it}. Tasks vary along four axes---episode length, observation modality, reward semantics, and termination---each over a range wide enough that no single representative collapses it. The abstraction that absorbs all of this is small: an environment exposes a \texttt{reset}/\texttt{step} interface, and a per-episode handle carries all transient state through the pipeline. The narrowness of this interface is what allows one asynchronous runtime to drive episodes of widely differing shape through the same workers. Section~\ref{sec:task-abstraction} develops the engineering choices behind it.

The second is the \textbf{resource abstraction}: the interface between the orchestrator and the resources whose cost and operational irregularity dominate evaluation. Such resources are compute-bound rather than async-bound, require multi-node deployment, and have failure modes whose timescales would dominate any orchestrator that hosted them directly. DeepInsight delegates each such resource class to its own \emph{control plane} and exposes it back through a narrow handle: ask for a handle, release it, and that is all the orchestrator does. Backend variance stays inside the planes; the orchestrator stays asynchronous and thin. Section~\ref{sec:resource-abstraction} develops the engineering choices behind it.

The third is the \textbf{result abstraction}: the interface between \emph{what the system records} and \emph{what it reports}. Every event the system produces is written into one structured record, addressed by a uniform identity tuple. Reported metrics are computed from this trace rather than retained in place of it; an aggregate localizes to a trace coordinate, and a new analysis enters as a new reader over the same trace rather than as a modification of the pipeline. Section~\ref{sec:result-abstraction} develops the engineering choices behind it.

\begin{figure}[t]
  \centering
  \includegraphics[width=\linewidth]{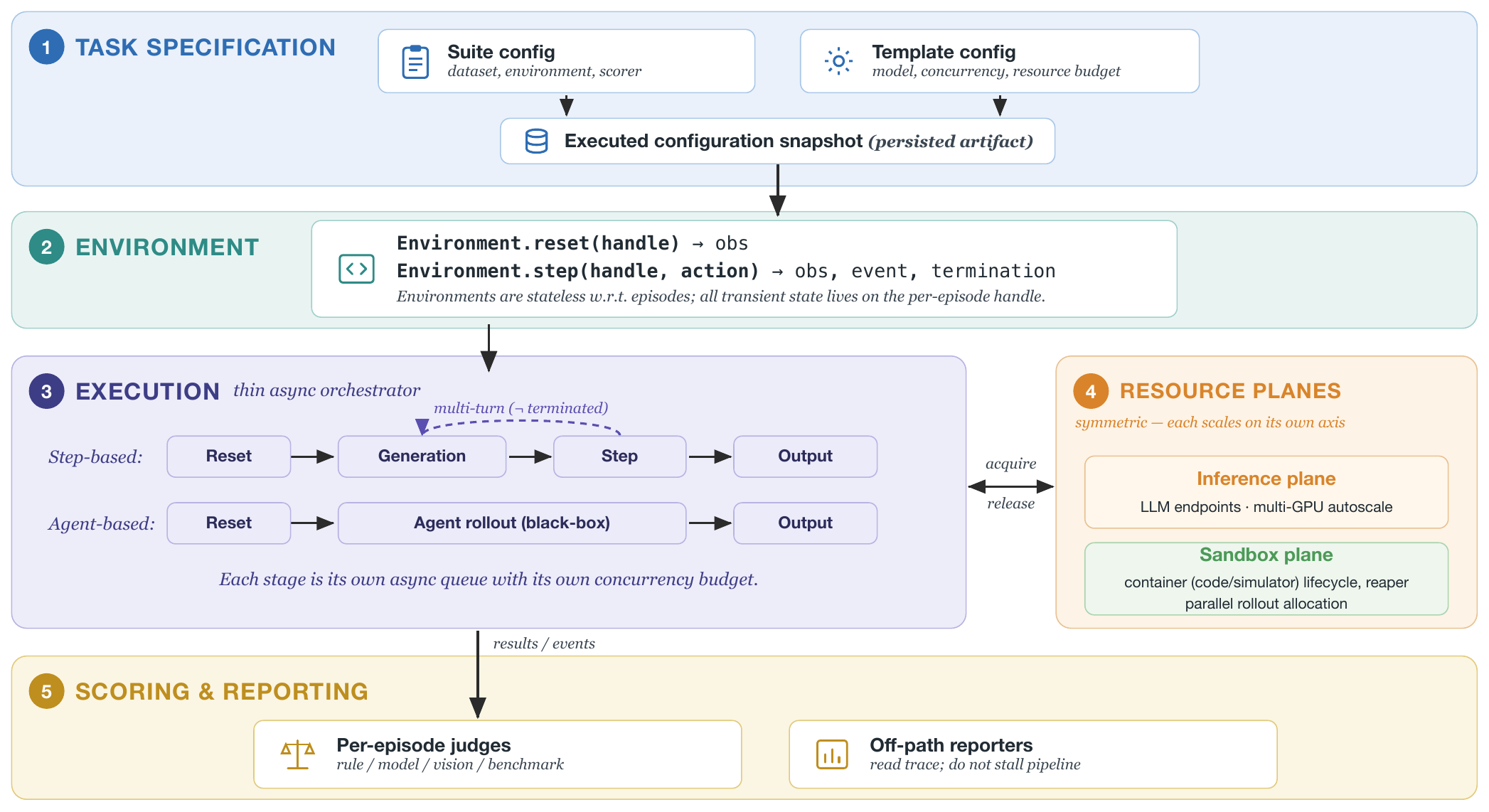}
  \caption{DeepInsight's architecture. Four pipeline stages (task specification, environment, execution, scoring \& reporting) run top to bottom, with the resource-plane bank side-mounted on execution. Execution acquires capacity from the bank via a shared \texttt{acquire}/\texttt{release} handle protocol and emits results directly to scoring; the two planes (inference, sandbox) are structurally identical from the orchestrator's perspective.}
  \label{fig:deepinsight-architecture}
\end{figure}

Figure~\ref{fig:deepinsight-architecture} shows the system that realizes these three abstractions. Four pipeline stages run top to bottom---task specification, environment, execution, and scoring \& reporting---and the resource-plane bank sits to the side of execution as a service rather than as a stage in the main flow. Three carrier mechanisms make the abstractions concrete: a \emph{per-episode handle} carries the task abstraction through the main pipeline, holding all transient episode state---the dataset sample, the conversation, the in-flight result, the current observation--action pair---so that environments and workers themselves remain stateless across episodes; \emph{plane handles} carry the resource abstraction across the execution-to-plane boundary, isolating execution from the operational machinery behind each plane; and a \emph{structured trace record} carries the result abstraction from the moment any subsystem writes an event to the moment scoring reads it. The main-flow arrow from execution to scoring delivers per-episode results and events, while the resource planes write their own lease and inference events directly into the trace.

The three abstractions are not parallel choices; they are causally linked. Driving operator-heterogeneous episodes on one runtime requires a task abstraction that hides episode shape from the workers driving them. A unified task driver then surfaces a heterogeneous resource bottleneck---LLM inference is throughput-bound, while sandboxed runtimes span I/O-bound code containers and compute-bound physics simulators---which the resource abstraction absorbs by detaching each class into its own control plane. The heterogeneous execution that follows produces heterogeneous events, which the result abstraction absorbs by collapsing them into one schema under one identity. Each step's output is the next step's input; the choice of three is not aesthetic.

``Single runtime'' here denotes four shared elements: one episode driver, one scheduler over per-stage worker pools, one resource-handle protocol that every plane implements, and one trace identity scheme under which all subsystems write. The inference and sandbox planes run as separate control planes---each with its own deploy lifecycle and failure modes---but they participate in evaluation only through the shared handle protocol and only by writing into the shared trace. A federation of per-segment harnesses linked by a top-level dispatcher would share none of these; the distinction between coordination and a shared runtime is structural, not nominal.

Together these three abstractions are the architecture's response to the spectrum-coverage gap diagnosed in Section~\ref{sec:related-work}. The joint adoption of all three---rather than any one in isolation---is what allows the infrastructure to span the spectrum across its full range, foundation-model end through whole-body-control end, on a single runtime.

\subsection{The Task Abstraction}
\label{sec:task-abstraction}

Three engineering choices realize the task abstraction: state lives on a per-episode handle, the executed configuration is the persisted artifact, and judgment is an axis orthogonal to the task. The patterns themselves---stateless servers with per-request state, configuration-snapshot replay, decoupling of metric from task---are not novel; the contribution is their composition under the constraint of spanning the spectrum.

\paragraph{State lives on the per-episode handle.}
A conventional pipeline stores per-task state on the environment itself---the dataset cursor, the in-progress conversation, the sandbox handle. This binds one environment to one episode: each instance becomes single-use, or fresh ones must be allocated per episode, which is expensive at scale and fragile around external resources. Under concurrency, residual state in shared environments makes which episodes interleave with which others a function of scheduling order.

DeepInsight inverts this. Environments are stateless with respect to episodes; all transient state---the dataset sample, the in-flight result, the conversation, the current observation--action pair---lives on the per-episode handle that workers carry through the pipeline. One environment instance can serve arbitrarily many concurrent episodes because every read and write goes through this handle. The structural payoff is at the runtime layer: a one-step QA decode and a several-hundred-step sandboxed agent rollout, despite using different environment implementations, share the same workers and the same async loop. Episode-bound state lives on the handle, not on the environment, so workers do not specialize to one shape or the other. This is the prerequisite for a task abstraction that crosses spectrum segments at all.

The interface that results is a two-method surface (Figure~\ref{fig:deepinsight-architecture}): \texttt{reset(handle)} returns an observation, and \texttt{step(handle, action)} returns an observation, a per-step event, and a termination class. Every environment in the deployed inventory---from a static-QA decoder to a sandbox-coupled coding rollout to a physics-bound humanoid simulator---implements exactly this surface. Episode-scoped state is read from and written to \texttt{handle}; the \texttt{Environment} object itself holds only configuration. The termination value carries the episode's exit class (success, failure mode, timeout); the event value carries the per-step record that the result abstraction (Section~\ref{sec:result-abstraction}) will later persist.

\paragraph{The executed configuration is the persisted artifact.}
If tasks are to be declared rather than written, then the form they actually execute in must itself survive the run. Task configurations in real pipelines fragment across multiple artifacts: a specification of what the task is, a specification of how to run it, override mechanisms, and inherited defaults. What actually executes is their merger, and is rarely captured in a form that survives the run.

DeepInsight separates declaration into two layers. A suite configuration pins what a task is---dataset, environment, scoring. A template configuration pins one run of it---model endpoints, concurrency, resource budgets. At startup the two merge with any per-field overrides into a single snapshot, dumped to disk and used for every replay. The snapshot captures every override actually applied, including those that vanish into command-line flags or environment defaults in less disciplined pipelines.

The same separation makes the declared form expressive enough for use directly. Because the suite alone declares what a task is, the dominant fraction of deployed benchmarks enters the system as configuration rather than code: one generic environment serves the majority of deployed suites, and bespoke environments appear only where reward generation falls outside declarative scoring, such as sandbox verifiers, tool-call validators, and simulator-based user models. The onboarding cost of a new benchmark, in the dominant case, is one configuration file. This is the mechanism through which the deployed inventory continues to grow: a new benchmark---at any point along the spectrum---is brought online as a declarative suite as long as its environment fits the carrier and its scoring fits the judge interface, without modifying the runtime.

\paragraph{Judgment is an axis orthogonal to the task.}
Many evaluation harnesses embed the scoring function in the task: the task computes its own metric. This couples two things that should evolve independently. When a scorer is upgraded---from exact to fuzzy match, from rule to model-based judge, from heuristic to in-environment verifier---the task's semantics shift with the metric. Two runs marked ``same task, different versions'' might differ because the model changed, or because the scoring did, with no clean way to tell.

DeepInsight makes judgment an orthogonal axis. Tasks define what is recorded---the trajectory, the termination class, the model's final output. Scorers are separately registered components that read the trace and emit rewards. The system carries four scorer families---rule-based, model-based, vision-specific, and benchmark-specific---sharing one interface: a trace in, a structured judgment out. A new judge applies across the corpus by substitution rather than per-task integration; the reported score is the cross product (task~$\times$~judge), each independently versioned.

Orthogonality here is interface-level, not absolute. A trajectory-analytic judge needs the environment to record the trajectory features it consumes; a model-based judge of an agentic rollout needs the runtime to persist the full conversation. The coupling runs through the trace schema---data-shape, not code path---and does not propagate into the scoring layer.

Together, the three structural choices let the task abstraction carry the full spectrum on a single runtime, and let extension along the spectrum reduce, in the dominant case, to configuration.

\subsection{The Resource Abstraction}
\label{sec:resource-abstraction}

Scaling an evaluation infrastructure looks deceptively like scaling any compute workload, until the heterogeneity of evaluation stages is taken seriously. Resources at the spectrum's foundation-model end are throughput-bound, those of its agentic middle are I/O-bound, and those of its whole-body-control end are physics-bound; each scales on its own axis. Three sources of scaling friction recur when this is ignored: stages with different resource profiles forced to share a concurrency budget, expensive backends hosted by the orchestrator process, and scaling knobs whose effects entangle across workloads. The resource abstraction addresses each.

\paragraph{Stages get independent concurrency budgets.}
Evaluation tasks decompose into stages that instantiate the resource profiles
introduced above. A generation step is GPU-bound, dominated by token decoding.
Sandboxed execution splits further: deploying a stateful sandbox is API-bound,
dominated by a multi-second wait on a remote control service; a code-test step
is I/O-bound and slow; and a lightweight verifier or deterministic scorer is
CPU-bound, fast, and cheap. Physics-simulation stages are compute-bound and
handled by the same sandbox plane. If these stages share one concurrency budget,
the slow stages monopolize workers while the cheap ones starve, and throughput
collapses to the rate of whichever stage is currently saturating the pool.

DeepInsight separates stages at a finer granularity than role alone: a CPU-bound step and an I/O-bound step are different queues even when they share the same step role. Each queue gets its own worker pool and concurrency budget, scaled to the bottleneck of its own profile. A slow sandbox step occupies only its own pool; cheap judge steps proceed in parallel through a separate one. The orchestrator becomes a flow controller across heterogeneous lanes rather than a single rate limiter, and the throughput at the spectrum's foundation-model end is no longer hostage to the queue characteristics of the spectrum's agentic middle.

\paragraph{Resources are delegated to symmetric control planes.}
The expensive resources of evaluation---language-model inference and sandboxed execution---bring more than slowness; they are operationally unruly. An inference replica can lag autoscale under load; a sandboxed runtime can leak resources between runs, or, when it hosts a physics simulator, drift in step rate. If the orchestrator hosts these resources, its async budget is consumed by their failure modes, and the infrastructure's throughput becomes a function of backend operational health.

DeepInsight's response was sketched in Section~\ref{sec:architecture-overview}: detach the two resource classes into their own control planes---inference and sandbox---and expose only narrow handles to the orchestrator. The detail that closes the scaling friction is what each plane does behind its handle. The inference plane scales inference capacity autonomously: multi-GPU, multi-node deployment with autoscaling under load. The sandbox plane manages the full lifecycle of sandboxed runtimes autonomously, from deployment under load through resource accounting to recovery from failure. The same plane carries workloads whose resource profiles differ widely---I/O-bound code containers used by tool-using agents, compute-bound physics simulators used by whole-body control evaluation, parallel rollouts that produce trajectory streams---because what unifies them at the infrastructure level is the lifecycle (allocate, lease, reap, restart), not the inner workload profile. To the orchestrator, both planes look the same: ask for a handle, release it, and that is all. Backend operational variance stays inside the planes; the orchestrator's async loop interacts only with the handle.

The symmetry across the two planes is the point. Each takes a request from a worker, returns a handle, and manages everything behind that handle autonomously. The same interface hides inference autoscaling, container deploy-storms, and parallel simulator allocation alike. Whatever runs behind the handle, the orchestrator sees the same shape.

\begin{figure}[t]
  \centering
  \begin{subfigure}[t]{0.49\linewidth}
    \centering
    \includegraphics[width=\linewidth]{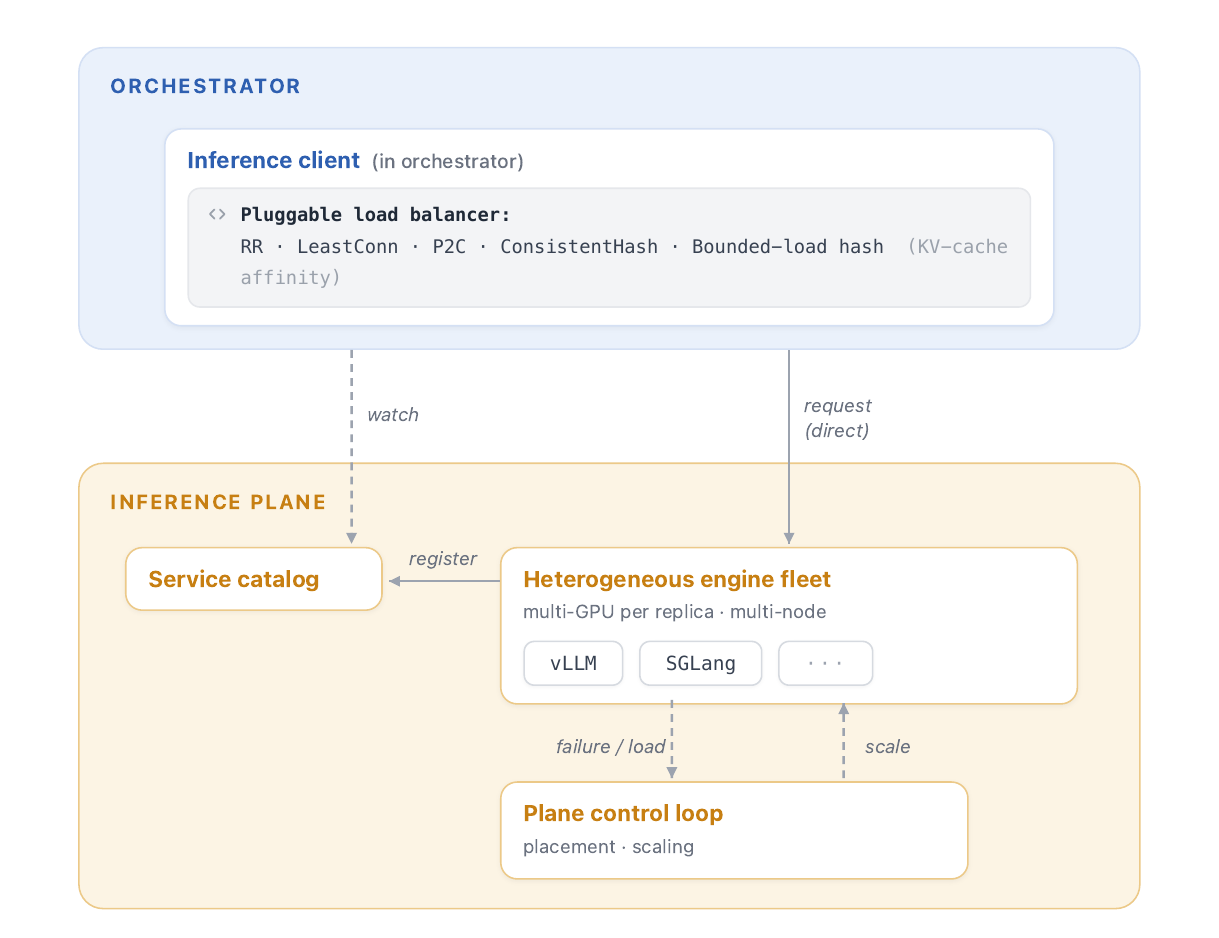}
    \caption{\textbf{Inside the inference plane.} A caller-side client with a pluggable load balancer watches a service catalog and routes requests directly to a heterogeneous engine fleet; engine-side telemetry feeds the plane's scaling loop.}
    \label{fig:inference-plane}
  \end{subfigure}
  \hfill
  \begin{subfigure}[t]{0.49\linewidth}
    \centering
    \includegraphics[width=\linewidth]{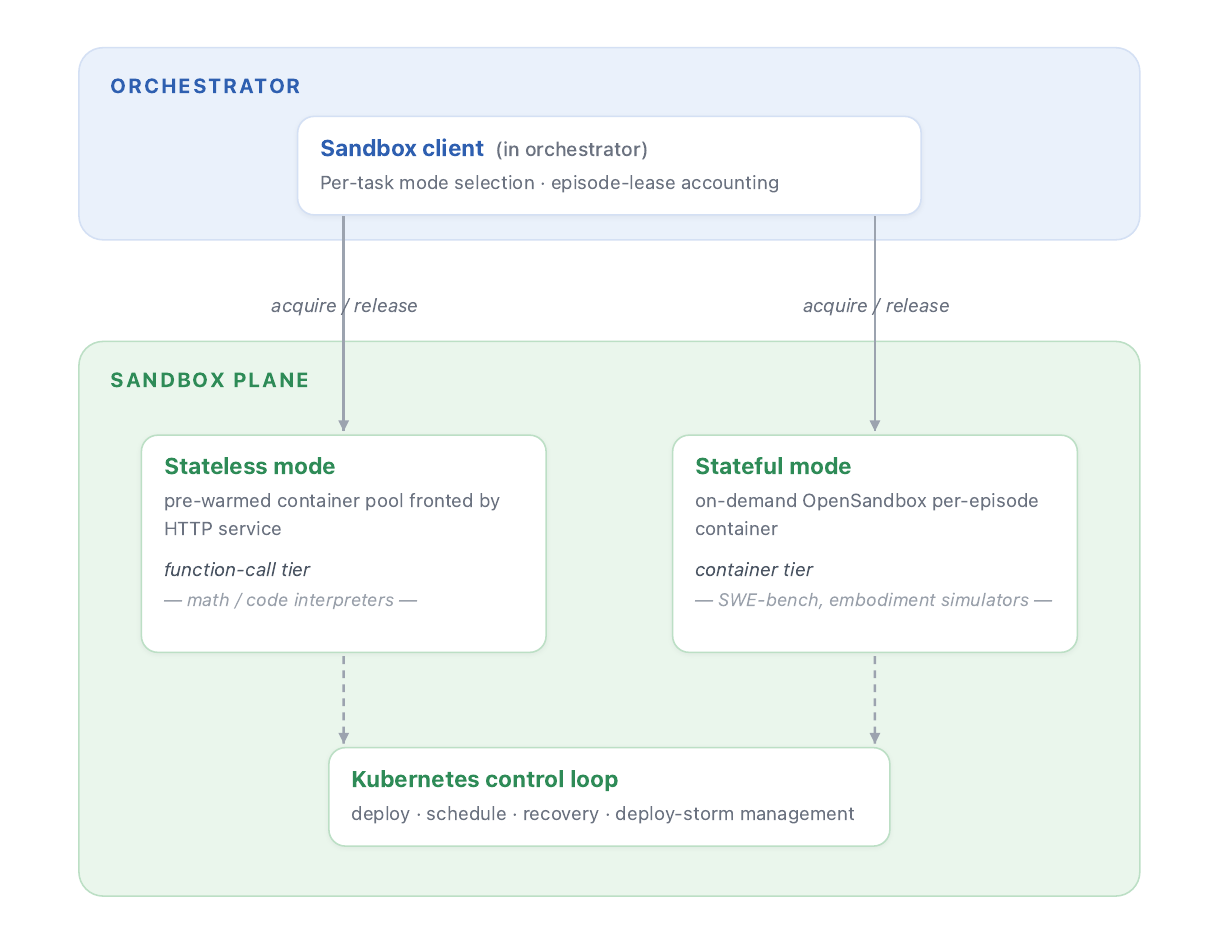}
    \caption{\textbf{Inside the sandbox plane.} Two deployment modes sit behind one handle: a pre-warmed pool fronted by HTTP for stateless tool calls, and an on-demand OpenSandbox provisioner that leases a per-episode container for stateful workloads. Lifecycle rides the underlying Kubernetes control loop.}
    \label{fig:sandbox-plane}
  \end{subfigure}
  \caption{What sits behind the two plane handles. Both planes expose the same \texttt{acquire}/\texttt{release} handle protocol upward; the contents below are the per-plane mechanisms that handle protocol hides.}
  \label{fig:resource-planes}
\end{figure}

\paragraph{Inside the inference plane.}
Behind the handle, the plane carries a heterogeneous engine fleet---vLLM, vLLM-Omni, SGLang---uniformly, so a benchmark expressed for one model is re-run against another by swapping a configuration field rather than rewriting the resource path. It places those engines across the GPU pool with parallelism shape fixed at plan time and admits new replicas into the live deploy without orchestrator restart. Service discovery is client-side: engines register their endpoints into a catalog that the caller-side transport watches, so requests reach the chosen replica without a gateway hop, and the load-balancing policy is free to condition on application-level routing keys---including a bounded-load consistent-hash policy that gives KV-cache affinity for prefix-cache hits with automatic spillover under hot keys. Engine-side telemetry---failure events and load---feeds the plane's scaling loop. Figure~\ref{fig:inference-plane} sketches what sits behind the handle.
\paragraph{Inside the sandbox plane.}
Sandboxed tasks vary in their state needs---and, with them, in deploy cost: a math benchmark's interpreter call is request-scoped and tolerates no multi-second deploy wait, while a SWE-bench rollout's filesystem state must persist across hundreds of agentic turns inside a per-episode container. The plane carries two deployment modes behind one handle to match. A long-running pool of pre-warmed containers fronted by an HTTP service serves the stateless case, eliminating the API-bound deploy wait that would otherwise dominate per-call latency. An on-demand provisioner leases a fresh OpenSandbox container per request, binds its lifetime to the episode handle of Section~\ref{sec:task-abstraction}, and reclaims it on release for the stateful case. The two modes occupy the function-call and container tiers of the four-tier isolation spectrum (function-call $\to$ container $\to$ microVM $\to$ fullVM) catalogued in recent industrial practice~\citep{deepseekv4}; higher-isolation tiers are reachable through the same handle but unneeded by current workloads. Lifecycle---deploy, schedule, recovery, deploy-storm management---rides the underlying Kubernetes control loop; the plane sits above this, selecting the mode, binding each lease to its episode, and accounting for it. Figure~\ref{fig:sandbox-plane} sketches the arrangement.

That shape is two methods: \texttt{acquire(kind, episode\_id, constraints)} returns a handle, and \texttt{release(handle)} returns nothing. The \texttt{kind} field discriminates inference and sandbox; \texttt{episode\_id} ties the lease to the per-episode handle of Section~\ref{sec:task-abstraction}, so that resource accounting joins on the same identity as the trace; \texttt{constraints} carries the per-class budget---model identifier and decoding parameters for inference; container image, resource cap, and (for physics-simulator workloads) simulator configuration and seed for sandbox. Beyond \texttt{acquire} and \texttt{release}, the orchestrator has no view into a plane's internals---autoscaling, deploy-storm management, and parallel-rollout allocation all live behind the handle.

\paragraph{Scaling knobs are orthogonal.}
Even with stages decoupled and backends delegated, a poorly factored scaling story can reintroduce coupling. If one knob controls multiple bottlenecks at once---for instance, a top-level concurrency setting that simultaneously caps worker count, connection-pool size, and sandbox deploys---then tuning becomes guesswork: raising the number to relieve one bottleneck pushes another past its limit.

DeepInsight exposes a scaling knob at each independent bottleneck: per-stage worker concurrency in the orchestrator, connection pool and replica count at the inference plane, and deploy / active-concurrency / parallel-rollout caps at the sandbox plane (set independently per workload class---code containers and physics simulators saturate on different axes). Each axis is independent. Tuning one does not deplete another: a workload that taxes sandbox capacity is tuned through sandbox knobs while leaving inference throughput intact.

Together, the three structural choices decouple the throughput bottlenecks across the spectrum, so that an optimization at one segment is not held hostage by the resource profile at another. The orchestrator stays asynchronous and thin; the planes scale autonomously beneath their handles; tuning adjusts independently along each axis. Section~\ref{sec:exp-efficiency} reports the throughput and stability consequences of these choices against open-source single-regime baselines along the spectrum.

\subsection{The Result Abstraction}
\label{sec:result-abstraction}

Evaluation produces a fan-out of event types---conversation turns, judge rationales, lease events, inference events, trajectory steps---that downstream analyses must read as one coherent trace. Three patterns recur in conventional pipelines that prevent this: heterogeneous result shapes from different subsystems, aggregates retained at the expense of the underlying trace, and reporting coupled into the evaluation pipeline. The result abstraction addresses each.

\paragraph{Events share one schema and one identity scheme.}
Different evaluation subsystems produce different kinds of events. A judge produces a reward, a rationale, and a reason. The runtime produces turn-by-turn dialogue, a termination class, and error counts. The inference plane produces inference events; the sandbox plane produces lease events and, where the sandboxed runtime is a physics simulator, per-step trajectory events with continuous state. In a conventional pipeline these would arrive in different shapes, in different stores, with different identifiers. Cross-source analysis---asking, for instance, whether the model's worst failures coincide with inference-plane saturation, or whether a manipulation policy's trajectory anomalies cluster with particular dataset partitions---becomes a multi-source data integration problem before it is a research question.

DeepInsight collapses this fan-out: every event from every subsystem writes into one structured record, addressed by a hierarchical identity that descends from the run to a position within an episode, and carrying its own causal lineage so that an event can be linked to the upstream event that produced it. The judge's reward, the runtime's dialogue, the sandbox plane's lease and trajectory events, and the inference plane's request event all share the same record shape and the same coordinate system. Cross-source analysis dissolves into a join on identity, and cross-layer causal analysis dissolves into a graph traversal over the parent pointer. The schema admits new event types---a perception failure from a new embodied environment, a fairness reading from a new judge family---through the same path, without retroactive migration of existing records.

\begin{figure}[t]
  \centering
  \includegraphics[width=0.95\linewidth]{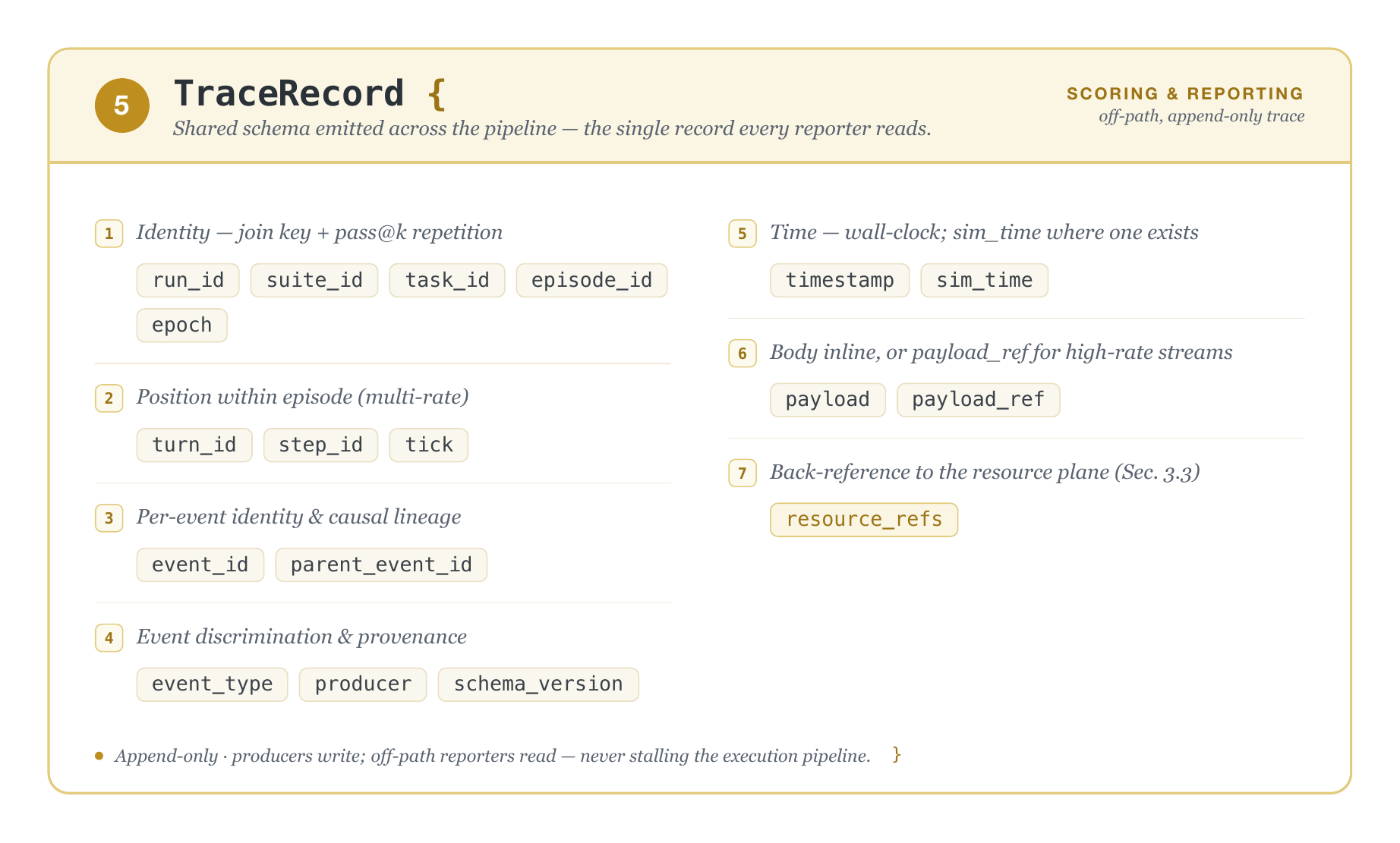}
  \caption{The \texttt{TraceRecord} schema. Identity, multi-rate position within an episode, per-event identity and causal lineage, event discrimination and provenance, time, payload, and the back-reference to the resource plane (Sec.~\ref{sec:resource-abstraction}) are grouped by role. The trace is append-only: every producer writes; off-path reporters read.}
  \label{fig:trace-record-schema}
\end{figure}

Figure~\ref{fig:trace-record-schema} sketches the schema, with fields grouped by role.

The schema separates three concerns. The first group is \emph{identity}: \texttt{(run\_id, suite\_id, task\_id, episode\_id, epoch)} is the join key, and every producer---runtime, scorer, inference plane, sandbox plane---writes the same tuple shape, so cross-source analysis is a join on any prefix rather than a multi-source reconciliation. The second group is \emph{position and lineage}: the multi-rate index \texttt{(turn\_id, step\_id, tick)} lets one trace carry a one-step QA decode, a multi-turn agentic rollout, and a whole-body-control rollout without changing shape, while \texttt{(event\_id, parent\_event\_id)} reduces cross-layer diagnosis to a graph traversal---a stabilizer regression at a tick walks back through its control decision, the policy generation, and the LLM call that produced the plan, on a single trace. The third group is \emph{content}: \texttt{event\_type} discriminates the producer's domain, \texttt{payload} is schema-versioned per type so a newly onboarded subsystem joins by extending the type registry rather than by migrating prior records, \texttt{payload\_ref} provides a by-reference escape hatch for the high-rate trajectory case, and \texttt{resource\_refs} links every event back to the inference endpoint or sandbox lease that produced it---so that a generation that coincided with an inference replica drop, a code-sandbox step that coincided with an eviction, or a physics-simulator trajectory anomaly tied to a specific lease is localizable on a single trace by a single join.

\paragraph{The trace is the unit of retention.}
Most evaluation pipelines retain the aggregate and discard the trace, or retain the trace transiently and discard it after aggregation. The implicit assumption is that aggregate scores are the product and that traces are incidental---useful at the moment of execution and discarded soon after. When a regression surfaces a week later, the trace that would explain it has been overwritten.

DeepInsight inverts the priority. The trace is the durable artifact: every event written under the schema above is persisted. Aggregate scores are computed from this trace, not retained in its place. A regression observed in aggregate can be traced back to its constituent events long after the run: the events themselves are still on disk. Drill-down becomes a read operation. Across spectrum segments this matters disproportionately: at the foundation-model end the aggregate hides individual generations; at the whole-body-control end it hides per-step physics state. In both cases the explanation of a degraded metric lives in the events, not in the number.

\paragraph{Reporting is decoupled from evaluation.}
Even when traces are retained, conventional pipelines often compute aggregates within the evaluation pipeline itself: each task's reporter runs as part of the pipeline, and modifying or extending the aggregation path requires changing the pipeline. The aggregate computed at pipeline exit is the only one available.

In DeepInsight, reporting consumes the trace as a downstream stage. Multiple reporters operate on the same trace within a run, off the hot path: an expensive aggregation does not stall evaluation, and adding a new reporter is a registration, not a pipeline modification. The reporters consume the same record format that was written by the upstream subsystems---the trace is uniformly addressed end-to-end, and reporters are simply its first consumers. The evaluation pipeline produces events; reporters produce numbers.

Together, the three structural choices let the events produced across the
spectrum accumulate into one trace that downstream analyses can read. A new
event type from a newly onboarded benchmark joins through the same schema; a new
analysis on existing or newly onboarded tasks joins as a new reader on the
trace. Section~\ref{sec:evaluation} evaluates metric fidelity and throughput
where peer references exist, while Section~\ref{sec:case-studies} uses the same
retained trace for cross-layer diagnosis.


%

\section{Evaluation}
\label{sec:evaluation}
\label{sec:experiments}   
\label{sec:exp-span}      
\label{sec:exp-efficiency}
\label{sec:exp-system2}   

System~2 is the only layer of the stack where DeepInsight competes with mature
open-source frameworks, and therefore the only one where we can \emph{measure}---
rather than merely demonstrate---what the infrastructure delivers. This section
establishes that production-grade depth along three axes: DeepInsight reproduces
reference scores at peer-framework fidelity, runs the same suites faster on a
single node, and scales near-linearly across nodes. The reach across the rest of
the stack, which no peer comparison can quantify, is the subject of
Section~\ref{sec:case-studies}. All numbers below come from the deployed
infrastructure in its production configuration, on a subset of the suites it runs
continuously---not a one-off benchmarking setup.

We compete against lm-evaluation-harness and Inspect~AI on text-only benchmarks,
and VLMEvalKit and lmms-eval on multimodal VQA. We proceed in three steps. First,
we show that DeepInsight reproduces reference scores, both against published
model-card numbers and against each peer framework running the same task on the
same model. Second, we show that it runs the same suites faster on a single node.
Third, we attribute the speedup to the resource abstraction of
Section~\ref{sec:resource-abstraction}.

All System~2 comparisons use the same hardware envelope, inference backend, and
benchmark protocol across frameworks. Each model is served through the same
vLLM configuration on a single 8$\times$A100 node, and each benchmark family
uses the official recommended sampling profile for that model. Model-based
scoring likewise uses a shared judge configuration wherever a judge or user
simulator is required. What differs by framework is only the scheduler surface:
concurrency is set to the highest-throughput setting appropriate to each
framework's own execution model, so the per-row readings below pair each
framework's score with the scheduler setting it was designed to use.

The primary model under test is Qwen3.6-27B---a dense 27B-parameter model with a
hybrid linear-and-full-attention stack. We also include one Qwen3-32B
text-alignment table under its official thinking protocol as a protocol-transfer
check; it is not used in the throughput claims. Tasks driven by an LLM judge or
user simulator---SimpleQA, $\tau$-bench, and the multimodal VQA suite---share the
same Qwen3-30B-A3B-Instruct-2507 judge configuration across all frameworks,
removing judge-model variance as a source of disagreement. SimpleQA follows the
short-form factuality benchmark of \citet{wei2024simpleqa}.

The only additional model family is the omni-modal setting in
Table~\ref{tab:system2-alignment-omni}, where we run
Qwen3-Omni-30B-A3B-Instruct under the same single-node vLLM-Omni configuration
and official recommended decoding profile. Among the peer frameworks, only
lmms-eval natively supports omni-modal evaluation, so Table~\ref{tab:system2-alignment-omni}
compares DeepInsight with lmms-eval.

\subsection{Accuracy and cross-framework alignment}
\label{sec:system2-alignment}

We compare DeepInsight against each peer framework on every task the peer
natively supports. The goal is alignment, not improvement: DeepInsight should
recover the published reference and peer-framework readings to within the
residual spread of the peer runs themselves. This fidelity check is the
precondition for the throughput comparison in Section~\ref{sec:system2-throughput}.

Tables~\ref{tab:system2-alignment-text-qwen36}--\ref{tab:system2-alignment-omni}
report the reference-anchored comparison on task families covered by at least one
peer framework. The two text tables are complementary: Table~\ref{tab:system2-alignment-text-qwen36}
reports Qwen3.6-27B production-model rows, while
Table~\ref{tab:system2-alignment-text-qwen3} adds Qwen3-32B official-protocol
rows, including additional math benchmarks and BFCL-v3. Table~\ref{tab:system2-alignment-mm}
extends the test to the 15 multimodal VQA rows shared by VLMEvalKit and
lmms-eval; Table~\ref{tab:system2-alignment-omni} covers the five omni-modal rows
shared with lmms-eval. Each row reports the benchmark's primary metric; dashes
mean the framework does not natively support the task, \textbf{bold} marks the
cell closest to Ref., and the final row counts closest-to-reference rows.

The text rows draw from MMLU-Pro~\citep{wang2024mmlupro},
MMLU-Redux~\citep{gema2025mmluredux}, SuperGPQA~\citep{du2025supergpqa},
GPQA-Diamond~\citep{rein2024gpqa}, C-Eval~\citep{huang2023ceval},
Humanity's Last Exam~\citep{phan2025hle}, MathArena competition
sets~\citep{balunovic2025matharena}, LiveCodeBench~\citep{jain2025livecodebench},
LiveBench~\citep{white2025livebench}, MATH/MATH-500~\citep{hendrycks2021math,lightman2024letsverify},
IFEval~\citep{zhou2023ifeval}, and BFCL~\citep{patil2025bfcl}. The multimodal
rows draw from MMMU~\citep{yue2024mmmu}, MMMU-Pro~\citep{yue2025mmmupro},
MathVista~\citep{lu2024mathvista}, DynaMath~\citep{zou2025dynamath},
BlindTest/VLMs-Are-Blind~\citep{rahmanzadehgervi2024vlmsblind},
MMBench~\citep{liu2024mmbench}, MMStar~\citep{chen2024mmstar},
RealWorldQA~\citep{xai2024realworldqa}, SimpleVQA~\citep{cheng2025simplevqa},
CharXiv~\citep{wang2024charxiv}, OCRBench~\citep{liu2024ocrbench},
CountBench~\citep{paiss2023countclip}, RefCOCO~\citep{yu2016refcoco},
ERQA~\citep{geminirobotics2025}, and Video-MME~\citep{fu2025videomme}. The
omni-modal rows use LibriSpeech~\citep{panayotov2015librispeech},
WeNetSpeech~\citep{zhang2022wenetspeech}, and
WorldSense~\citep{hong2025worldsense}; their Ref.\ values come from the
Qwen3-Omni technical report~\citep{xu2025qwen3omni}.

\begin{table}[t]
\centering
\scriptsize
\setlength{\tabcolsep}{4.5pt}
\renewcommand{\arraystretch}{1.16}
\caption{Reference-anchored text alignment on Qwen3.6-27B under the production
thinking profile. Entries are primary-metric percentages; \textbf{bold} marks
closest to Ref., and the final row counts closest-to-reference rows.}
\label{tab:system2-alignment-text-qwen36}
\begin{tabularx}{\linewidth}{>{\raggedright\arraybackslash}p{0.18\linewidth}
                              >{\raggedright\arraybackslash}X
                              >{\centering\arraybackslash}p{0.08\linewidth}
                              >{\centering\arraybackslash}p{0.13\linewidth}
                              >{\centering\arraybackslash}p{0.13\linewidth}
                              >{\centering\arraybackslash}p{0.15\linewidth}}
\toprule
Class & Benchmark & Ref. & lm-eval & Inspect AI & DeepInsight \\
\midrule
\multirow{6}{*}{Knowledge \& reasoning}
  & MMLU-Pro                  & 86.2 & 83.14\,$\pm$\,0.04 & \textbf{86.49\,$\pm$\,0.07} & 85.44\,$\pm$\,0.07 \\
  & MMLU-Redux                & 93.5 & 94.34\,$\pm$\,0.13 & ---  & \textbf{92.99\,$\pm$\,0.01} \\
  & SuperGPQA                 & 66.0 & ---  & ---  & \textbf{66.25\,$\pm$\,0.05} \\
  & GPQA-Diamond              & 87.8 & 84.85\,$\pm$\,1.39 & \textbf{86.03\,$\pm$\,1.68} & 83.38\,$\pm$\,2.10 \\
  & C-Eval                    & 91.4 & 90.81\,$\pm$\,0.10 & ---  & \textbf{91.50\,$\pm$\,0.03} \\
  & HLE                       & 24.0 & ---  & \textbf{23.95\,$\pm$\,0.59} & 25.17\,$\pm$\,0.80 \\
\midrule
Math
  & HMMT Feb 2025             & 93.8 & ---  & ---  & \textbf{93.33\,$\pm$\,2.04} \\
\midrule
Code
  & LiveCodeBench v6          & 83.9 & --- & --- & \textbf{75.61\,$\pm$\,0.49} \\
\midrule
\multicolumn{2}{l}{\textbf{Closest to Ref.}}
  & --- & 0/8 & 3/8 & \textbf{5/8} \\
\bottomrule
\end{tabularx}
\end{table}

\begin{table}[t]
\centering
\scriptsize
\setlength{\tabcolsep}{4.5pt}
\renewcommand{\arraystretch}{1.16}
\caption{Reference-anchored text alignment on Qwen3-32B under the official
thinking profile. Rows complement Table~\ref{tab:system2-alignment-text-qwen36};
entries are primary-metric percentages; \textbf{bold} marks closest to Ref.,
and the final row counts closest-to-reference rows.}
\label{tab:system2-alignment-text-qwen3}
\begin{tabularx}{\linewidth}{>{\raggedright\arraybackslash}p{0.18\linewidth}
                              >{\raggedright\arraybackslash}X
                              >{\centering\arraybackslash}p{0.08\linewidth}
                              >{\centering\arraybackslash}p{0.13\linewidth}
                              >{\centering\arraybackslash}p{0.13\linewidth}
                              >{\centering\arraybackslash}p{0.15\linewidth}}
\toprule
Class & Benchmark & Ref. & lm-eval & Inspect AI & DeepInsight \\
\midrule
\multirow{4}{*}{Knowledge \& reasoning}
  & MMLU-Redux                & 90.9 & \textbf{90.79\,$\pm$\,0.08} & --- & 89.29\,$\pm$\,0.06 \\
  & GPQA-Diamond              & 68.4 & 60.61\,$\pm$\,1.92 & \textbf{63.89\,$\pm$\,0.03} & 60.61\,$\pm$\,3.47 \\
  & C-Eval                    & 87.3 & \textbf{89.38\,$\pm$\,0.84} & --- & 84.59\,$\pm$\,0.09 \\
  & LiveBench                 & 74.9 & --- & 65.24\,$\pm$\,0.21 & \textbf{74.25\,$\pm$\,0.25} \\
\midrule
\multirow{3}{*}{Math}
  & MATH-500                  & 97.2 & \textbf{96.80\,$\pm$\,0.79} & --- & 93.50\,$\pm$\,0.30 \\
  & AIME-2024                 & 81.4 & 72.71\,$\pm$\,4.72 & 73.54\,$\pm$\,2.76 & \textbf{76.04\,$\pm$\,4.44} \\
  & AIME-2025                 & 72.9 & 56.25\,$\pm$\,5.46 & \textbf{59.79\,$\pm$\,7.12} & 58.75\,$\pm$\,6.11 \\
\midrule
Instruction \& QA
  & IFEval                    & 85.0 & 83.36\,$\pm$\,1.60 & 81.34\,$\pm$\,0.78 & \textbf{83.45\,$\pm$\,0.09} \\
\midrule
Tool use (BFCL)
  & BFCL-v3                   & 70.3 & --- & 43.37\,$\pm$\,0.02 & \textbf{50.63\,$\pm$\,0.89} \\
\midrule
\multicolumn{2}{l}{\textbf{Closest to Ref.}}
  & --- & 3/9 & 2/9 & \textbf{4/9} \\
\bottomrule
\end{tabularx}
\end{table}

\begin{table}[t]
\centering
\scriptsize
\setlength{\tabcolsep}{3.5pt}
\renewcommand{\arraystretch}{1.20}
\caption{Multimodal VQA alignment on the 15 Qwen3.6-27B benchmarks shared by all
three frameworks. Entries are mean\,$\pm$\,std over 3 runs; \textbf{bold} marks
closest to Ref., and the final row counts closest-to-reference rows with ties
split.}
\label{tab:system2-alignment-mm}
\begin{tabularx}{\linewidth}{>{\raggedright\arraybackslash}p{0.14\linewidth}
                              >{\raggedright\arraybackslash}X
                              >{\centering\arraybackslash}p{0.06\linewidth}
                              >{\centering\arraybackslash}p{0.16\linewidth}
                              >{\centering\arraybackslash}p{0.16\linewidth}
                              >{\centering\arraybackslash}p{0.16\linewidth}}
\toprule
Class & Benchmark & Ref. & VLMEvalKit & lmms-eval & DeepInsight \\
\midrule
\multirow{5}{*}{STEM \& Puzzle}
  & MMMU (DEV\_VAL)              & 82.9 & 82.04\,$\pm$\,0.67 & 71.75\,$\pm$\,0.54 & \textbf{83.14\,$\pm$\,0.49} \\
  & MMMU-Pro (10c CoT)           & 75.8 & 75.90\,$\pm$\,0.40 & 72.99\,$\pm$\,0.62 & \textbf{75.80\,$\pm$\,0.53} \\
  & MathVista-mini               & 87.4 & \textbf{87.50\,$\pm$\,0.44} & 85.70\,$\pm$\,0.53 & 87.90\,$\pm$\,0.36 \\
  & DynaMath                     & 85.6 & 84.52\,$\pm$\,0.42 & 70.14\,$\pm$\,0.05 & \textbf{86.59\,$\pm$\,0.30} \\
  & VlmsAreBlind                 & 97.0 & 90.23\,$\pm$\,0.05 & 88.59\,$\pm$\,0.28 & \textbf{90.26\,$\pm$\,0.23} \\
\midrule
\multirow{4}{*}{General VQA}
  & MMBench EN-DEV               & 92.3 & 86.57\,$\pm$\,0.35 & 87.28\,$\pm$\,0.23 & \textbf{92.74\,$\pm$\,0.13} \\
  & MMStar                       & 81.4 & 78.64\,$\pm$\,0.50 & \textbf{80.90\,$\pm$\,0.79} & 80.49\,$\pm$\,0.31 \\
  & RealWorldQA                  & 84.1 & 83.40\,$\pm$\,0.34 & 83.22\,$\pm$\,0.84 & \textbf{83.83\,$\pm$\,0.59} \\
  & SimpleVQA                    & 56.1 & \textbf{57.01\,$\pm$\,0.30} & 51.13\,$\pm$\,0.52 & 57.52\,$\pm$\,0.08 \\
\midrule
\multirow{2}{*}{Document}
  & CharXiv (reasoning)          & 78.4 & 78.57\,$\pm$\,0.78 & 72.97\,$\pm$\,0.50 & \textbf{78.47\,$\pm$\,0.67} \\
  & OCRBench                     & 89.4 & \textbf{88.43\,$\pm$\,0.25} & 87.87\,$\pm$\,0.21 & 88.40\,$\pm$\,1.25 \\
\midrule
\multirow{3}{*}{Spatial}
  & CountBench                   & 97.8 & \textbf{97.40\,$\pm$\,0.31} & \textbf{97.40\,$\pm$\,0.31} & 96.92\,$\pm$\,0.90 \\
  & RefCOCO                      & 92.5 & \textbf{92.19\,$\pm$\,0.16} & 92.89\,$\pm$\,0.12 & 91.63\,$\pm$\,0.53 \\
  & ERQA                         & 62.5 & \textbf{58.83\,$\pm$\,1.23} & \textbf{58.83\,$\pm$\,1.16} & 58.25\,$\pm$\,1.39 \\
\midrule
Video
  & Video-MME (64f)              & 87.7 & 67.63\,$\pm$\,0.35 & 65.56\,$\pm$\,0.17 & \textbf{74.88\,$\pm$\,0.92} \\
\midrule
\multicolumn{2}{l}{\textbf{Closest to Ref.}}
  & --- & 5.0/15 & 2.0/15 & \textbf{8.0/15} \\
\bottomrule
\end{tabularx}
\end{table}

\begin{table}[t]
\centering
\scriptsize
\setlength{\tabcolsep}{4pt}
\renewcommand{\arraystretch}{1.20}
\caption{Omni-modal alignment on the five Qwen3-Omni benchmarks shared with
lmms-eval. Entries are mean\,$\pm$\,std over 3 runs; \textbf{bold} marks closest
to Ref., and the final row counts closest-to-reference rows.}
\label{tab:system2-alignment-omni}
\begin{tabularx}{\linewidth}{>{\raggedright\arraybackslash}p{0.20\linewidth}
                              >{\raggedright\arraybackslash}X
                              >{\centering\arraybackslash}p{0.10\linewidth}
                              >{\centering\arraybackslash}p{0.08\linewidth}
                              >{\centering\arraybackslash}p{0.15\linewidth}
                              >{\centering\arraybackslash}p{0.15\linewidth}}
\toprule
Benchmark & Modality & Metric & Ref. & lmms-eval & DeepInsight \\
\midrule
LibriSpeech-clean   & audio$\to$text      & WER $\downarrow$ & 1.22 & 1.438\,$\pm$\,0.004 & \textbf{1.412\,$\pm$\,0.007} \\
LibriSpeech-other   & audio$\to$text      & WER $\downarrow$ & 2.48 & 2.625\,$\pm$\,0.014 & \textbf{2.566\,$\pm$\,0.020} \\
WeNetSpeech-meeting & audio$\to$text (ZH) & CER $\downarrow$ & 5.89 & \textbf{5.906\,$\pm$\,0.008} & 5.806\,$\pm$\,0.010 \\
WeNetSpeech-net     & audio$\to$text (ZH) & CER $\downarrow$ & 4.69 & 4.768\,$\pm$\,0.004 & \textbf{4.728\,$\pm$\,0.011} \\
WorldSense          & video+audio MCQ     & acc $\uparrow$   & 54.0 & 49.21\,$\pm$\,0.06 & \textbf{53.03\,$\pm$\,0.18} \\
\midrule
\multicolumn{3}{l}{\textbf{Closest to Ref.}}
  & --- & 1/5 & \textbf{4/5} \\
\bottomrule
\end{tabularx}
\end{table}

\paragraph{Cross-framework alignment.}
Across the four alignment tables, DeepInsight stays on the reference scale while
covering the broadest set of modalities. It is closest to Ref.\ on the largest
number of rows in every table: $5/8$ on Qwen3.6 text, $4/9$ on Qwen3-32B text,
$8.0/15$ on shared multimodal VQA, and $4/5$ on omni-modal evaluation. This is
the fidelity condition needed for the throughput comparison: the systems being
timed are producing comparable scores, not trading accuracy for speed.

The remaining offsets are protocol-level rather than infrastructure-level.
LiveCodeBench depends on an unpublished reference sandbox/timeout; Video-MME is
compared against a with-subtitles reference while all open runs use no subtitles.
These cases bound the alignment evidence instead of
weakening it: where the protocol is matched, DeepInsight reproduces
reference-scale readings across text, multimodal VQA, and omni-modal suites.

\subsection{End-to-end throughput}
\label{sec:system2-throughput}

Given the alignment established above, we compare wall-clock on each peer's
native task surface. Open-source peers do not share a common task surface:
lm-eval covers
short-episode text benchmarks, Inspect~AI extends to tool use and agentic
rollouts, and VLMEvalKit and lmms-eval cover multimodal VQA. They also differ
in \emph{how} they schedule that work, along the two axes DeepInsight's engine
targets (Section~\ref{sec:resource-abstraction}). lm-eval, its multimodal fork
lmms-eval, and VLMEvalKit are phase-structured and batch-oriented: they
generate over the dataset in synchronous passes and score afterward, with the
heterogeneous stages sharing a single global concurrency setting. Inspect~AI is
the exception on the first axis---it drives samples through an \texttt{asyncio}
event loop with adaptive model-connection concurrency, so generation is already
pipelined continuously rather than dispatched in batches---but on the second
axis it behaves like the rest, routing generation, tool/sandbox execution, and
scoring through one shared concurrency budget rather than giving each stage an
independent pool. None of the four runs multi-node. We therefore make one
pairwise comparison per peer rather than fixing a single suite for all peers.
For each peer, the dataset list is exactly what that peer natively supports on
Qwen3.6-27B; the same list is submitted to DeepInsight, and both frameworks
schedule it natively on a single 8$\times$A100 node under the configuration of
Section~\ref{sec:system2-alignment}. Wall-clock is the elapsed time of the full
suite.

\begin{center}
  \includegraphics[width=0.8\linewidth]{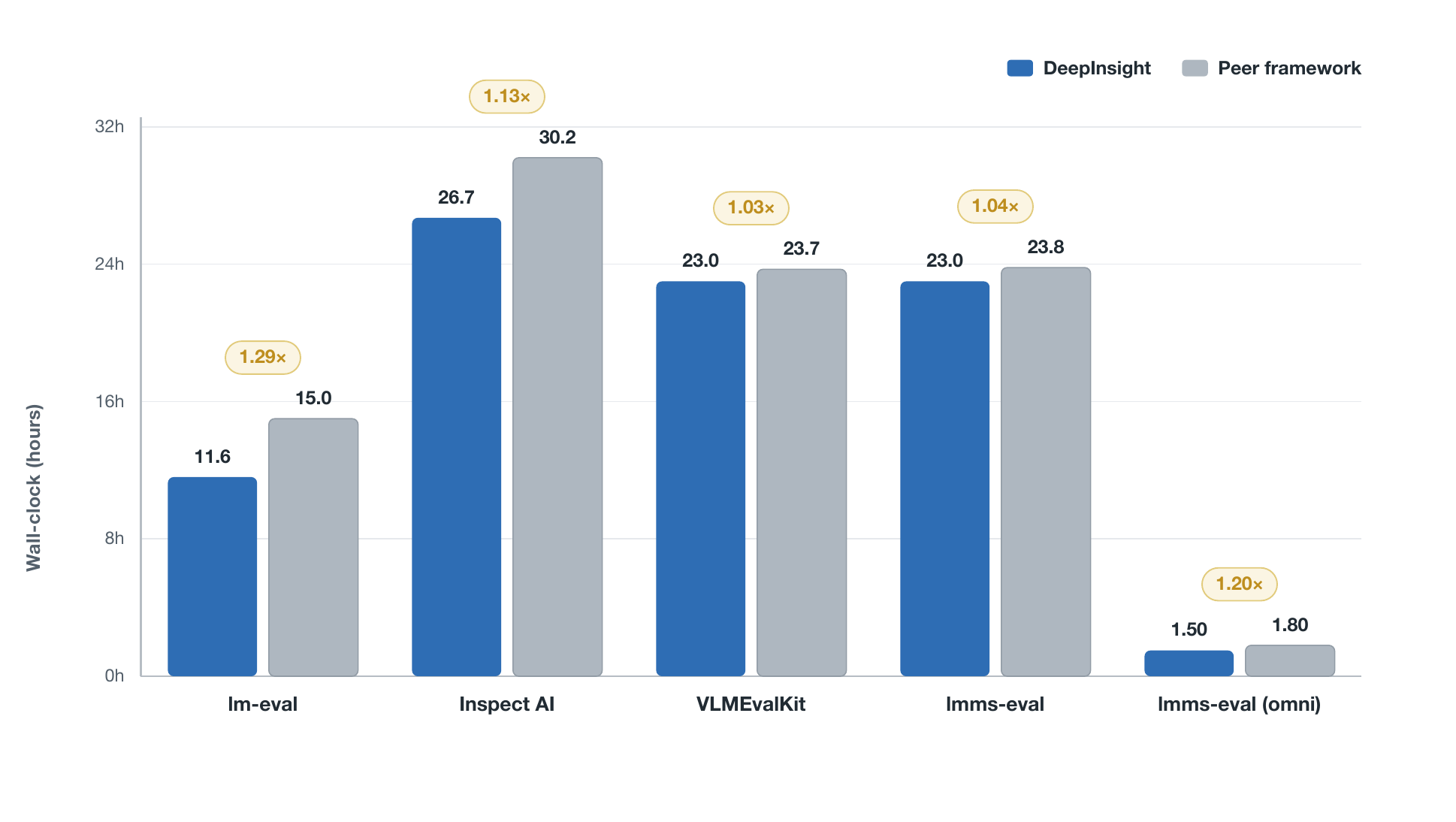}
  \captionof{figure}{End-to-end suite-level wall-clock comparison, single
  8$\times$A100 node. Each pair of bars is one peer comparison on that peer's
  native dataset list. Wall-clock is the elapsed time of the full suite,
  reported in hours. Speedup is the peer wall-clock divided by DeepInsight's;
  $>1$ favors DeepInsight. The omni row uses Qwen3-Omni-30B-A3B-Instruct; all
  other rows use Qwen3.6-27B.}
  \label{fig:system2-speedup}
\end{center}

The Qwen3-32B protocol check is fidelity-only and is excluded from the
wall-clock comparison. All comparisons are single-node because the peer frameworks do not support multi-node execution;
DeepInsight's own multi-node scaling has no peer baseline and is reported
separately in Section~\ref{sec:system2-throughput-source}.

The speedup tracks two things: how much idle time a workload's output-length
variance leaves for the scheduler to reclaim, and which of DeepInsight's engine
mechanisms the peer lacks. Multimodal VQA is prefill-bound---image encoding
dominates per-episode cost and decoder outputs are short and roughly
uniform---so there is little to reclaim, and the speedup over VLMEvalKit and
lmms-eval compresses to $1.03$--$1.04\times$; the omni suite, mixing long ASR
transcripts with short-form MCQ, sits at $1.20\times$. The lm-eval list is where
the batch-oriented discipline costs most---its math and extended-thinking rows
are long and length-variable, so the slowest episode in each synchronous pass
holds back the rest, and asynchronous pipelining reclaims exactly those idle
slots for $1.29\times$. Inspect~AI carries a comparably long agentic and
tool-use workload yet shows a \emph{smaller} $1.13\times$, precisely because it
is already asynchronous and does not pay that batch-barrier cost; what remains
against it traces to its stages still sharing one concurrency budget rather than
to pipelining. Section~\ref{sec:system2-throughput-source} isolates both
mechanisms directly---asynchronous pipelining on AIME-2024, where output-length
variance is maximal, and stage decoupling on a workload whose stage profiles
diverge sharply.


\subsection{Where the throughput comes from}
\label{sec:system2-throughput-source}

A speedup number does not explain itself, and the cross-framework comparison of
Section~\ref{sec:system2-throughput} cannot by itself isolate cause: a peer
differs from DeepInsight in implementation as well as architecture. We
therefore reproduce each peer's scheduling discipline \emph{inside} DeepInsight
and re-measure on a fixed System~2 workload, disabling one structural choice
from Section~\ref{sec:resource-abstraction} at a time while holding the model,
sampling, judge, and hardware fixed. Two of the three choices map directly onto
the architectural axes just described: forcing a synchronous batch barrier
reproduces the batch-oriented discipline of lm-eval, lmms-eval, and VLMEvalKit,
and collapsing the stages into one shared pool reproduces the single-budget
discipline common to all four peers (Inspect~AI included). Because only the
disabled choice varies, any speedup it recovers is attributable to that choice
rather than to incidental differences---so these internal ablations are what
license reading the cross-framework trend as a consequence of architecture. The
third choice, horizontal scaling, has no peer baseline, since none of the peers
run multi-node.

\begin{figure}[t]
  \centering
  \includegraphics[width=0.95\linewidth]{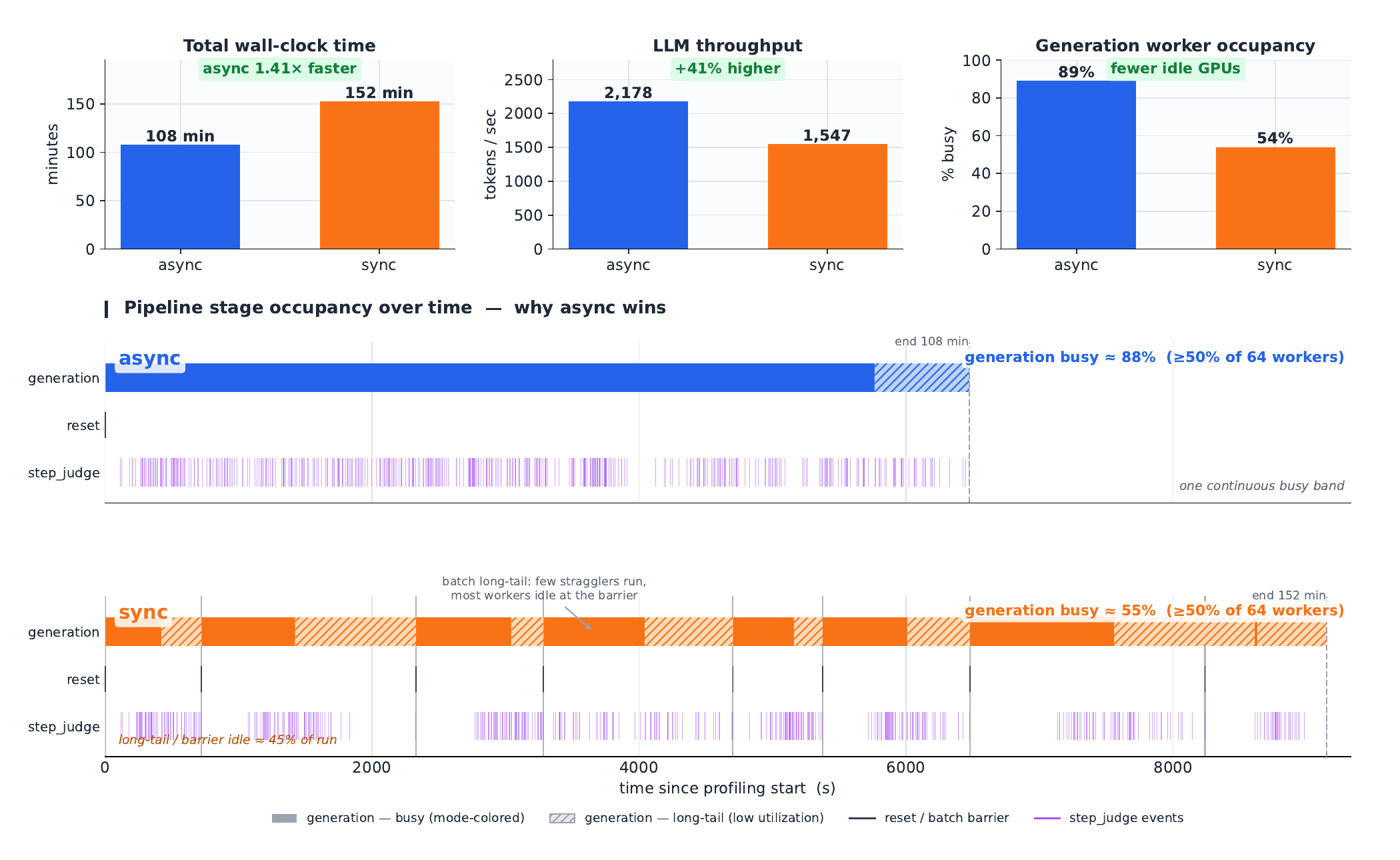}
  \caption{Asynchronous pipelining ablation on AIME-2024. \textbf{Top:}
  wall-clock and LLM throughput under async refill vs.\ synchronous batch
  barriers. \textbf{Bottom:} generation occupancy over wall-clock time.}
  \label{fig:async-ablation}
\end{figure}

The first choice is asynchronous pipelining. We isolate its effect on
AIME-2024 ($30$ problems $\times 16$ repeats $= 480$ episodes, the standard
mean-at-$16$ protocol for this benchmark), varying only the engine's
scheduling discipline: async refills from the queue continuously, whereas sync imposes a
$64$-episode batch barrier. Figure~\ref{fig:async-ablation} reports the
result. Asynchronous pipelining alone cuts wall-clock from $152$ to $108$
minutes---a $1.41\times$ speedup---while preserving accuracy to within the
run-to-run variance of the $16$-sample mean ($94.79$ vs.\ $95.63\%$). The
per-stage occupancy trace makes the mechanism concrete: under the batch
barrier, the slowest episode in each group of $64$ stretches the batch past
its median and leaves the generation pool idle through the long tail of every
batch (generation occupancy $54\%$), whereas async frees each fast episode's
slot the moment it finishes, holding generation at $89\%$ occupancy in one
continuous band and lifting LLM throughput from $1{,}547$ to $2{,}178$
tokens/s ($+41\%$).

\begin{figure}[t]
  \centering
  \includegraphics[width=0.93\linewidth]{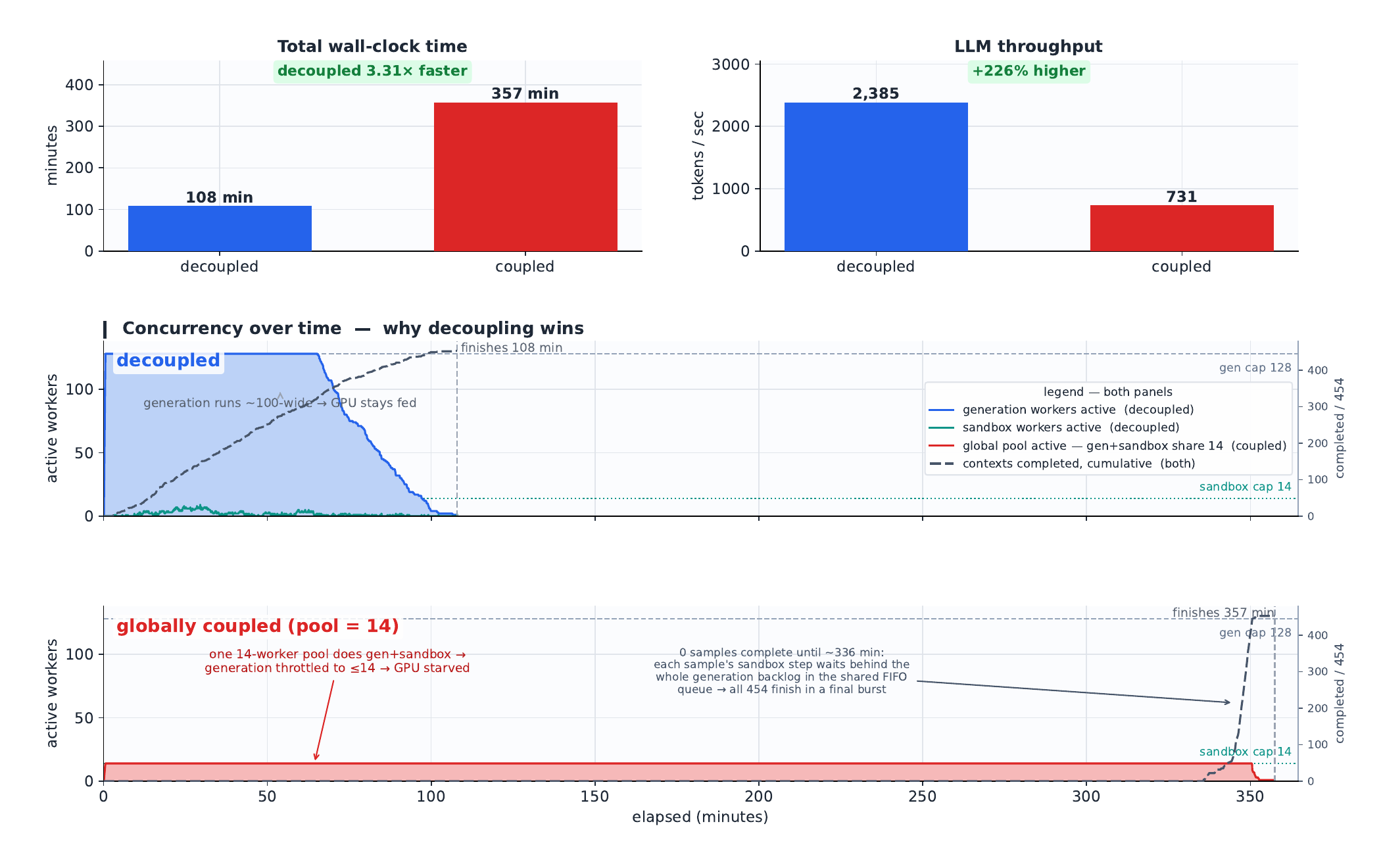}
  \caption{Stage-decoupling ablation on LiveCodeBench~v6. \textbf{Top:}
  wall-clock and LLM throughput under decoupled stage pools vs.\ a
  sized-to-sandbox-cap coupled lower bound, not a peer default configuration.
  \textbf{Bottom:} generation, sandbox, and completion concurrency over
  wall-clock time.}
  \label{fig:decouple-ablation}
\end{figure}

The second choice is stage decoupling. We isolate its effect on
LiveCodeBench~v6 ($454$ samples), whose stages have the sharpest profile
mismatch in the System~2 suite: generation is GPU-bound and runs at
concurrency $128$ across two eval instances, while sandbox code-execution is
I/O-bound and capped at $14$ concurrent containers by the sandbox plane. Two
conditions vary only the orchestrator's scheduling discipline. The
\emph{decoupled} condition runs each stage on its own queue and worker pool.
The \emph{globally coupled} condition collapses the steady-state stages into
a single global pool, which must be sized to the most constrained stage to
avoid backpressure---here $14$ workers, the sandbox cap. This is the
single-shared-budget regime every peer operates in, Inspect~AI included, but
sized to the sandbox cap as a coupled lower-bound ablation rather than as any
peer's default configuration. Heterogeneous stages contend for one concurrency
setting instead of each receiving its own pool. Figure~\ref{fig:decouple-ablation}
reports the result. Under this sized-to-sandbox-cap coupling, stage decoupling
yields a $3.31\times$ wall-clock speedup ($1$h$48$m vs.\ $5$h$57$m). The
coupled pool is busy $98\%$ of the time, but that occupancy is spent on the
wrong resource: with only $14$ shared slots, generation is throttled from its
$128$-way budget down to $14$, starving the GPU and collapsing LLM throughput
from $2{,}385$ to $731$ tokens/s. The decoupled orchestrator instead runs
sandbox at its $14$-way cap and generation at its full $128$-way budget
concurrently. Coupling carries a second penalty, visible in the completion
trace: because each sample's sandbox step queues behind the entire generation
backlog in the shared FIFO, no sample finishes until ${\sim}336$ minutes in,
after which all $454$ complete in a single final burst---so coupling also
forfeits any incremental results. Pass@$1$ differs by $\sim 3$~pp between
conditions ($75.55$ vs.\ $72.69$), within the single-pass binomial
uncertainty at $454$ items.

\begin{figure}[!htbp]
  \centering
  \begin{tikzpicture}
  \begin{axis}[
    width=0.62\linewidth,
    height=0.35\linewidth,
    xlabel={Number of nodes},
    ylabel={Speedup vs.\ 1 node},
    xmin=0.5, xmax=4.5,
    ymin=0.5, ymax=4.5,
    xtick={1, 2, 4},
    ytick={1, 2, 3, 4},
    tick align=outside,
    tick pos=left,
    grid=both,
    grid style={draw=black!8, line width=0.2pt},
    major grid style={draw=black!15, line width=0.3pt},
    legend style={
      draw=none, fill=none,
      at={(0.04,0.96)}, anchor=north west,
      font=\small, row sep=1pt,
    },
    legend cell align={left},
  ]
  \addplot+[
    draw=black!55,
    dashed,
    line width=0.6pt,
    mark=none,
    forget plot,
  ] coordinates {(0.5,0.5) (4.5,4.5)};
  \addplot+[
    draw=black!55,
    dashed,
    line width=0.6pt,
    mark=none,
  ] coordinates {(1,1) (2,2) (4,4)};
  \addlegendentry{Ideal linear}

  \addplot+[
    draw=PolicyPoint,
    line width=1pt,
    mark=*,
    mark size=2.5pt,
    mark options={draw=PolicyPoint, fill=PolicyPoint},
  ] coordinates {(1,1.00) (2,2.03) (4,4.00)};
  \addlegendentry{DeepInsight (measured)}

  \node[anchor=north west, font=\scriptsize, text=PolicyPoint, xshift=2pt]
    at (axis cs:1, 1.00) {$1.00\times$};
  \node[anchor=north west, font=\scriptsize, text=PolicyPoint, xshift=3pt, yshift=-2pt]
    at (axis cs:2, 2.03) {$2.03\times$};
  \node[anchor=east, font=\scriptsize, text=PolicyPoint, xshift=-3pt, yshift=8pt]
    at (axis cs:4, 4.00) {$4.00\times$};

  \end{axis}
  \end{tikzpicture}
  \caption{Horizontal scaling on the $27$-suite System~2 workload. Measured
  speedup is shown against the ideal linear reference across $1$, $2$, and
  $4$ nodes.}
  \label{fig:scaling-curve}
\end{figure}
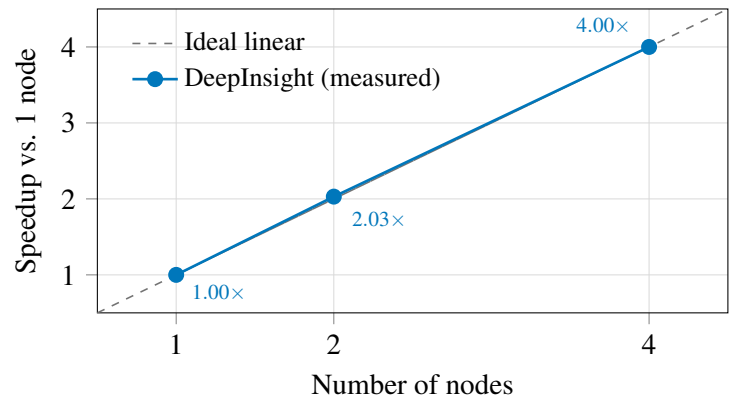

The third choice is horizontal scaling. We run the same $27$-suite System~2
workload at $1$, $2$, and $4$ nodes, with the
eval-instance count scaling with hardware ($2$/$4$/$8$ instances at TP$=4$
each). Figure~\ref{fig:scaling-curve} reports the result. Wall-clock halves
at each doubling---$80$h$55$m at $1$ node, $39$h$53$m at $2$ nodes, and
$20$h$14$m at $4$ nodes---yielding a $2.03\times$ speedup across the first
doubling, $1.97\times$ across the second, and $4.00\times$ across the full
$1 \to 4$ range, each within $\sim 1.5\%$ of linear; failure rates stay below
$0.1\%$ across all three runs, and the shared judge configuration is never the
bottleneck. Capacity grows with hardware without re-tuning the orchestrator,
which is the structural property the resource abstraction targets.

\FloatBarrier

\section{Case Studies across the Stack}
\label{sec:case-studies}

Where peer frameworks exist, the previous section measured DeepInsight
competitively. The rest of the stack has no such peers: no orchestrator spans
the visuomotor middle or the whole-body-control end, so here we exhibit what a
peer comparison cannot. The three case studies that follow form a single arc of
increasing reach. System~1 extends the substrate into closed-loop simulation and
subjective evaluation (Section~\ref{sec:exp-system1}); System~0 turns
trajectory-analytic evaluation into a release decision
(Section~\ref{sec:exp-system0}); and the full-system study composes all three
layers under one trace and localizes the failures that surface only after
composition (Section~\ref{sec:exp-full-system}). Each step reaches a regime no
existing framework touches; together they show the whole stack carried on one
runtime.

These are demonstrations of coverage, not competitive benchmarks. The System~1
and System~0 results report what the substrate can drive and how new evaluations
are onboarded; the quantitative depth of those two layers is left to future work.

\subsection{System~1: Reaching Closed-Loop Simulation and Subjective Evaluation}
\label{sec:exp-system1}

\begin{figure}[!htbp]
  \centering
  \includegraphics[width=0.82\linewidth]{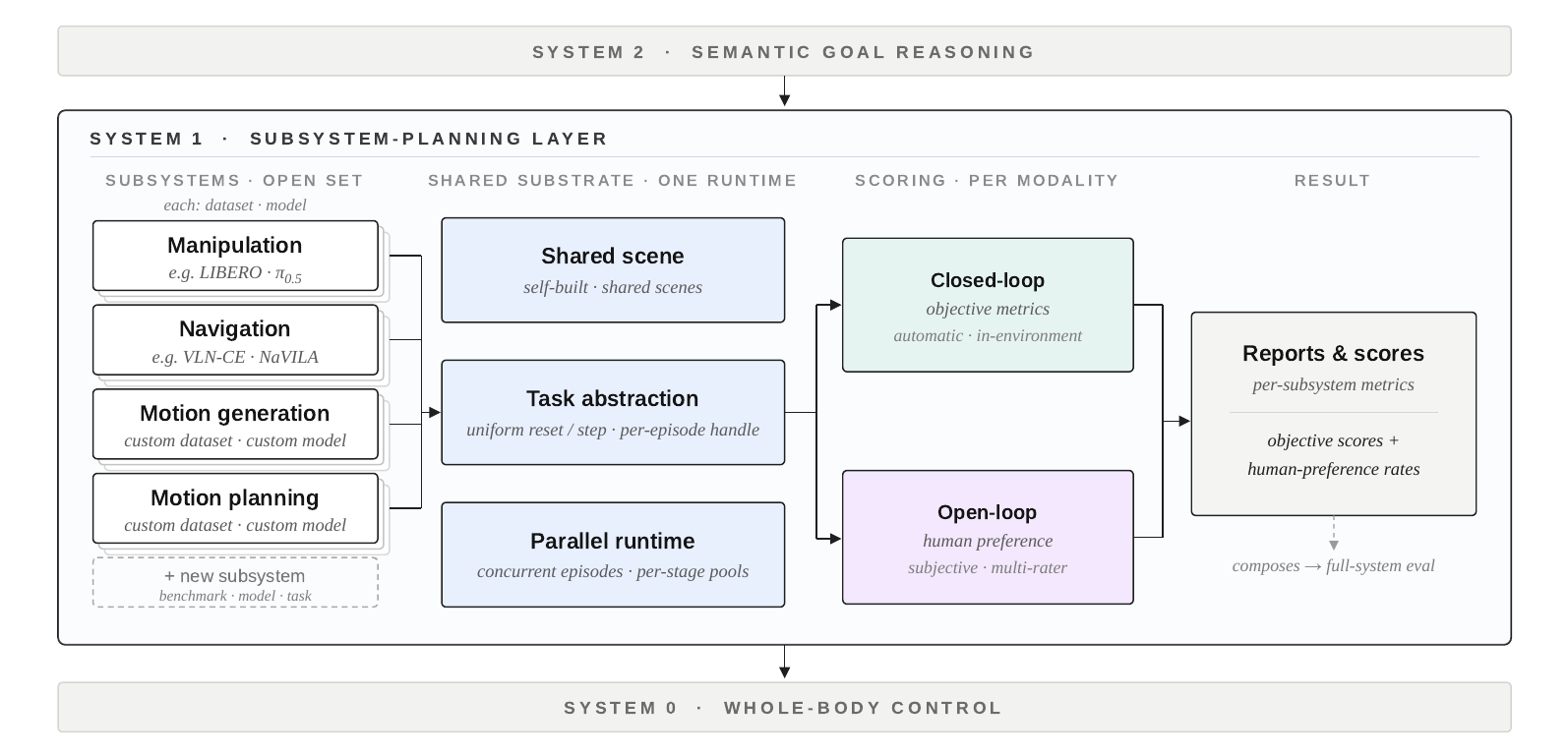}
  \caption{System~1 subsystem-planning evaluation on the unified infrastructure.
  Heterogeneous subsystem benchmarks share one simulation, execution, trace, and
  reporting path; benchmark-specific logic stays in adapters and scorers.}
  \label{fig:exp-s1-architecture}
\end{figure}

System~1 is the subsystem-planning layer between System~2's reasoning and
System~0's control---an open set of navigation, manipulation, and motion
subsystems. Its value as a case study is that it extends the evaluation
infrastructure into a regime the previous section does not touch: closed-loop
physics simulation and subjective human-preference evaluation, both supported by
a unified execution, trace, and reporting infrastructure for heterogeneous
System~1 subsystems. What we demonstrate is the evaluation coverage and
extensibility enabled by the unified closed-loop simulation infrastructure---not
exact score reproduction under the original benchmark environments.
Figure~\ref{fig:exp-s1-architecture} situates these subsystems on this unified
infrastructure.

\paragraph{Closed-loop coverage across subsystems.}
System~1 subsystems, such as navigation, manipulation, and motion generation,
are instantiated on DeepInsight's unified closed-loop simulation infrastructure
and shared execution/reporting path. Baseline examples include VLN-CE-style
navigation~\citep{krantz2020navgraph}, LIBERO-style
manipulation~\citep{liu2023libero}, and additional internally constructed tasks
with custom scenes and metrics. For each benchmark,
DeepInsight preserves the protocol elements that define the evaluation---suites,
splits, policies, success criteria, and metrics---while binding them to common
simulation, resource, trace, and reporting infrastructure. The point is not exact score
reproduction under a new simulator, but that structurally different System~1
evaluations coexist within one evaluation infrastructure without a separate
harness for each subsystem.

\paragraph{Subjective evaluation as a complementary modality.}
For System~1 outputs that interact with users through language or motion,
objective task or trajectory metrics are often insufficient, and human
preference provides a direct complementary evaluation signal. We use
audio-conditioned motion generation as an example: two anonymized model outputs
are compared on 20 audio clips, with ten blind raters making pairwise judgments
in four categories---A preferred, B preferred, both good, or both poor---across
scene/content match, motion completeness and smoothness, perceived safety, and
expressive style. DeepInsight records these judgments in the same result schema as
closed-loop evaluations, allowing subjective and objective evaluations to be
reported through the same infrastructure. Table~\ref{tab:exp-s1-motion}
(appendix) summarizes the categorical preference results.

\paragraph{Cross-benchmark evaluation extensibility.}
System~1 extensibility comes from localizing benchmark-specific assumptions to
well-defined adapter boundaries while keeping the underlying infrastructure
fixed. A new benchmark binds its scene content, raw data format, episode
specification, model interface, and metrics to the corresponding adapters and
reducers; the simulation, execution, model-interface, and reporting abstractions
are reused. For either a new public benchmark or an internal benchmark,
onboarding therefore requires implementing the benchmark-specific scene/asset
binding, task adapter, model binding, and evaluation reducers, rather than
changing the evaluation runtime itself.
Table~\ref{tab:exp-s1-extensibility} (appendix) summarizes where shared
infrastructure ends and benchmark-specific binding begins. System~1
extensibility is therefore not a new harness per benchmark, but the same
infrastructure with benchmark-specific bindings---a prerequisite for the
cross-layer evaluation in Section~\ref{sec:exp-full-system}.

\FloatBarrier

\subsection{System~0: From Aggregate Ranking to Release Decision}
\label{sec:exp-system0}

System~0 evaluates the whole-body controller (WBC) that makes upstream plans
physically executable. Here the case study makes a methodological point peer
benchmarks cannot: at this layer evaluation is not a number but a release
decision, and the decision is made on the trace. Routine training produces many
WBC policies, and the best aggregate row may still hide local contact, posture,
or joint-dynamics failures that block deployment. DeepInsight therefore runs a
two-stage, trace-backed workflow---summarized in
Figure~\ref{fig:system0-release-workflow}---a mechanical policy-set screen,
then a behavior-level diagnostic on the candidate the screen nominates.

\begin{center}
\begin{minipage}{\linewidth}
  \centering
  \includegraphics[width=0.86\linewidth]{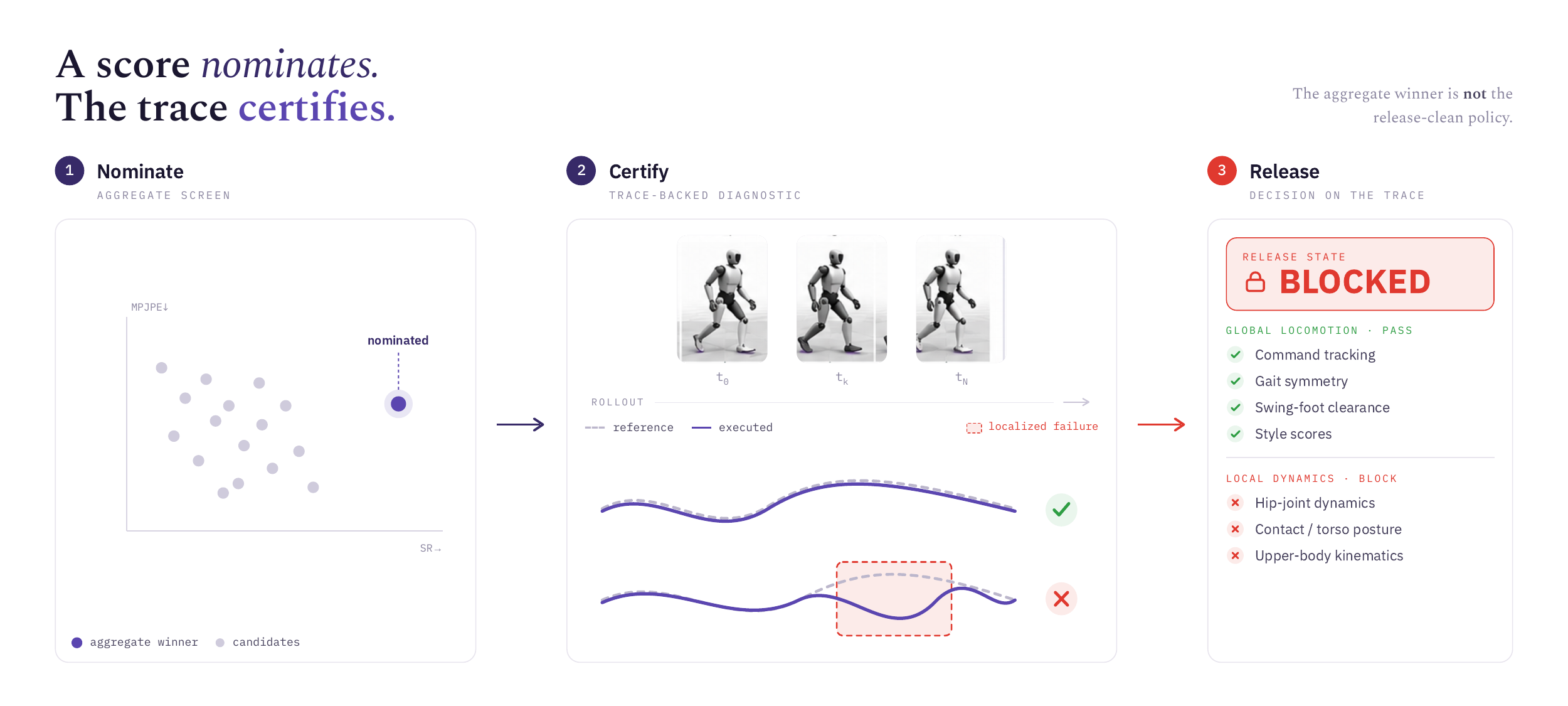}
  \captionof{figure}{Trace-backed System~0 release workflow. A policy-set score nominates
  the aggregate winner, but the selected policy is then certified against
  per-tick trajectory evidence on the shared trace.}
  \label{fig:system0-release-workflow}
\end{minipage}
\end{center}

\paragraph{Stage 1: a controlled aggregate screen.}
Every candidate faces the same robot, motions, rollout settings, and success
thresholds, so the screen is a controlled comparison rather than a set of curated
examples. For each policy DeepInsight exports a model-pool row with success rate
(SR) and mean per-joint position error (MPJPE), and the SR--MPJPE plane
(Figure~\ref{fig:system0-policy-plane}) makes the trade-off visible. By that
aggregate criterion, \textsc{WBC-RC-01} is the best policy in the run. The
leaderboard excerpt behind the plane and representative stills from the
auditable per-clip videos retained by the screen are given in the appendix
(Table~\ref{tab:system0-policy-set}, Figure~\ref{fig:system0-screening-frames}).

\begin{center}
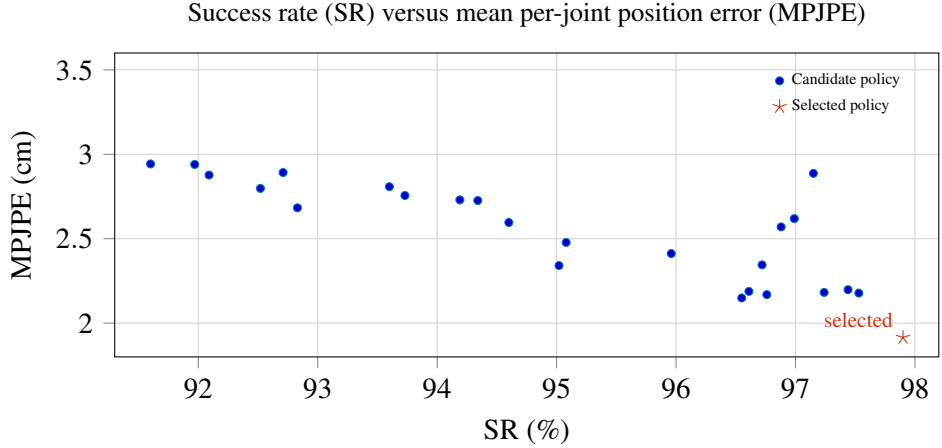

\begin{minipage}{\linewidth}
\centering
\vspace{-0.6em}
\begin{tikzpicture}
\begin{axis}[
  width=0.78\linewidth,
  height=0.35\linewidth,
  xmin=91.3, xmax=98.2,
  ymin=1.80, ymax=3.60,
  title={Success rate (SR) versus mean per-joint position error (MPJPE)},
  title style={font=\small, align=center},
  xlabel={SR (\%)},
  ylabel={MPJPE (cm)},
  tick align=outside,
  tick pos=left,
  grid=both,
  grid style={draw=black!10, line width=0.2pt},
  major grid style={draw=black!18, line width=0.3pt},
  legend style={draw=none, fill=none, at={(0.97,0.97)}, anchor=north east, font=\tiny},
  legend cell align={left},
]
\addplot+[
  only marks,
  mark=*,
  mark size=1.5pt,
  draw=PolicyPoint,
  fill=PolicyPoint!65,
] table[row sep=\\] {
sr mpjpe\\
97.53 2.178\\
97.44 2.198\\
97.24 2.182\\
97.15 2.887\\
96.99 2.619\\
96.88 2.570\\
96.76 2.169\\
96.72 2.345\\
96.61 2.188\\
96.55 2.149\\
95.96 2.412\\
95.08 2.478\\
95.02 2.341\\
94.60 2.596\\
94.34 2.726\\
94.19 2.730\\
93.73 2.756\\
93.60 2.808\\
92.83 2.683\\
92.71 2.892\\
92.52 2.798\\
92.09 2.877\\
91.97 2.940\\
91.60 2.943\\
};
\addlegendentry{Candidate policy}

\addplot+[
  only marks,
  mark=star,
  mark size=2.8pt,
  draw=SelectedPolicy,
  fill=SelectedPolicy,
] coordinates {(97.90,1.915)};
\addlegendentry{Selected policy}

\node[anchor=south east, font=\scriptsize, text=SelectedPolicy]
  at (axis cs:97.90,1.915) {selected};
\end{axis}
\end{tikzpicture}
\vspace{-0.5em}
\captionof{figure}{System~0 policy-set screening in the success-rate--tracking-error plane. Each point is one candidate WBC policy; the selected policy is highlighted.}
\label{fig:system0-policy-plane}
\vspace{-0.4em}
\end{minipage}
\end{center}

\paragraph{Stage 2: the aggregate winner, blocked on the trace.}
Winning the plane is necessary but not sufficient. \textsc{WBC-RC-01} is then put
through a behavior-level diagnostic checklist---command tracking, gait symmetry,
swing-foot clearance, style scores, hip-joint dynamics, contact attitude,
torso posture, upper-body kinematics---whose acceptance states are computed from registered
trajectory statistics, not visual inspection. The result is the contrast that
motivates the whole workflow (Table~\ref{tab:system0-mujoco-diagnostics}): the
aggregate winner passes the global locomotion checks for command tracking, gait
symmetry, swing-foot clearance, and style scores, yet fails the localized
checks for hip dynamics, contact and torso posture, and upper-body range of motion. The
diagnostic-metric schema and representative rollout video stills are in the appendix
(Table~\ref{tab:system0-diagnostic-schema}, Figure~\ref{fig:system0-mujoco-frames}).

\begin{table}[!htbp]
\centering
\vspace{0.75em}
\small
\setlength{\tabcolsep}{4pt}
\renewcommand{\arraystretch}{0.96}
\caption{Behavior-level diagnostic checklist for the selected policy.
Acceptance states are computed from registered trajectory statistics rather
than visual inspection.}
\label{tab:system0-mujoco-diagnostics}
\begin{tabularx}{\textwidth}{>{\raggedright\arraybackslash}p{0.27\textwidth}>{\raggedright\arraybackslash}X>{\raggedright\arraybackslash}p{0.17\textwidth}}
\toprule
Behavioral dimension & Diagnostic statistic & Acceptance state \\
\midrule
Command tracking & Linear-velocity and yaw-rate tracking residuals & Pass \\
Gait symmetry & Left--right step-length asymmetry & Pass \\
Swing-foot clearance & Mean foot clearance and bilateral clearance imbalance & Pass \\
Style scores & Forward, backward, turn, and rotate imitation scores & Pass \\
Hip-joint dynamics & Mean hip-pitch angular velocity & Fail \\
Contact-attitude stability & Touchdown foot-pitch angle; torso pitch angle & Fail \\
Upper-body kinematics & Elbow- and shoulder-pitch range of motion & Fail \\
\bottomrule
\end{tabularx}
\renewcommand{\arraystretch}{1.0}
\vspace{0.75em}
\end{table}

This is a judgment an aggregate-only pipeline cannot produce: it depends on
per-tick trajectory state being retained and queryable on one trace, so that a
release-blocking local failure stays localizable behind an otherwise winning
score. DeepInsight therefore does not stop System~0 evaluation at policy ranking;
it converts the screen, leaderboard, diagnostics, and pass/fail checks into a
single release-facing evidence chain, exposing why an aggregate winner may still
not be ready for deployment.
\FloatBarrier

\subsection{Full-System Validation: Cross-Layer Failure Localization}
\label{sec:exp-full-system}

Real-world robotic tasks are inherently compositional: a robot must combine
reasoning, tool use, skill execution, and physical control within one episode.
The preceding case studies evaluate each layer in isolation. They are necessary
but not sufficient for measuring the final task performance of the integrated
embodied system: System~2 evaluation can assess high-level reasoning without
verifying executable tool use; System~1 evaluation can measure local skills
without testing orchestration across tool calls; and System~0 evaluation can
expose stability without determining whether that stability supports task
completion. We therefore include a full-system case study to evaluate what
isolated benchmarks cannot: the task-level performance of the integrated
System~2--1--0 system and the failures that emerge only after the layers are
composed.
This study uses one representative composed task family to instantiate
integrated evaluation and cross-layer failure localization on a single trace;
broader coverage across task families is left to future work.

\paragraph{Task and protocol.}
We use a representative vehicle-guide task to evaluate the integrated
System~2--1--0 system in a showroom-like scene. Each episode contains a robot, a
user agent, and one or more vehicles, with their poses randomized within the
showroom. The robot receives a single task prompt: ``Find the user, greet them,
ask what help is needed, and then complete the requested assistance.'' The user
agent may respond freely; in the evaluated task instances, the interaction
enters the vehicle-guide branch when the user asks, ``Please introduce the car in
front of me.'' The robot must then resolve the deictic reference, reach a
presentation pose near the referenced vehicle, and deliver a spoken presentation
with accompanying motion.

The evaluation focuses on both the completed task and the system behavior that
produces it. We measure not only whether the vehicle-guide episode succeeds
end-to-end, but also how task understanding, user interaction, navigation, state
handoff, content grounding, presentation delivery, and physical execution
contribute to the final outcome.

Figure~\ref{fig:full-system-timeline} illustrates one representative episode
trace for this task. It shows how semantic analysis, navigation, user
interaction, speech delivery, presentation motion, low-level control events, and
the final judgment are recorded under the same episode identity.

\begin{center}
\begin{minipage}{\linewidth}
  \centering
  \resizebox{0.90\linewidth}{!}{%
  \begin{tikzpicture}[x=0.011cm,y=-0.011cm,font=\sffamily,
      nodebox/.style={draw,line width=0.8pt,rounded corners=4pt,align=center,
        font=\fontsize{7.2}{8.4}\selectfont,text width=1.6cm,inner xsep=3pt,inner ysep=5pt,
        minimum height=0.95cm,fill=white,anchor=center,
        execute at begin node={\hyphenpenalty=10000\exhyphenpenalty=10000\relax}},
      s2box/.style={nodebox,draw=LocS2,fill=CleS2bg,text=LocInk},
      s1box/.style={nodebox,draw=LocS1,fill=CleS1bg,text=LocInk},
      s0box/.style={nodebox,draw=LocS0,fill=CleS0bg,text=LocS0},
      urbox/.style={nodebox,draw=CleUR,fill=CleURbg,text=CleUR},
      lanelab/.style={nodebox,font=\fontsize{7.5}{8.8}\selectfont\bfseries,text width=1.2cm,inner xsep=3pt},
      thread/.style={CleThread,line width=1.5pt,rounded corners=5pt,line cap=round,
        -{Triangle[length=5pt,width=4pt]}},
      s0wire/.style={LocS0,line width=1.2pt,line cap=round,
        -{Triangle[length=4.2pt,width=3.6pt]}},
      userwire/.style={CleUR,line width=1.2pt,dash pattern=on 4pt off 3pt,rounded corners=5pt,
        line cap=round,-{Triangle[length=4.2pt,width=3.6pt]}}]
    \pgfsetlayers{cle-bg,cle-wires,main}

    \begin{pgfonlayer}{cle-bg}
      \fill[white,rounded corners=10pt] (10,140) rectangle (1470,806);
      \draw[CleS2bd,line width=1.2pt,rounded corners=10pt] (10,140) rectangle (1470,806);
      \draw[CleHair,line width=0.7pt] (10,320) -- (1470,320);
      \draw[CleHair,line width=0.7pt] (10,480) -- (1470,480);
      \draw[CleHair,line width=0.7pt] (10,638) -- (1470,638);
    \end{pgfonlayer}

    \node[anchor=east,font=\footnotesize,text=LocInk] at (1470,118)
      {\clesw{LocS2}{}~System 2\quad\clesw{LocS1}{}~System 1\quad%
       \clesw{LocS0}{}~System 0\quad\clesw{CleUR}{}~User};

    \node[lanelab,draw=LocS2,fill=CleS2bg,text=LocS2] at (112,235) {System 2};
    \node[lanelab,draw=LocS1,fill=CleS1bg,text=LocS1] at (112,400) {System 1};
    \node[lanelab,draw=LocS0,fill=CleS0bg,text=LocS0] at (112,559) {System 0};
    \node[lanelab,draw=CleUR,fill=CleURbg,text=CleUR] at (112,715) {User};

    \node[s2box,text width=1.7cm] (task) at (310,235) {Task decomposition};
    \node[s2box] (dialogue)  at (615,235) {User dialogue};
    \node[s2box] (resolve)   at (850,235) {Request grounding};
    \node[s2box] (knowledge) at (1160,235) {Knowledge retrieval};
    \node[s1box] (navuser)   at (310,400) {Navigate to user};
    \node[s1box] (greeting)  at (510,400) {Greeting};
    \node[s1box] (navveh)    at (970,400) {Navigate to vehicle};
    \node[s1box] (vehintro)  at (1160,400) {Speech delivery};
    \node[s1box] (presmot)   at (1350,400) {Presentation motion};
    \node[s0box] (loc1)      at (310,559) {Locomotion};
    \node[s0box] (wbc)       at (510,559) {Whole-body control};
    \node[s0box] (loc2)      at (970,559) {Locomotion};
    \node[s0box] (motgen)    at (1350,559) {Whole-body control};
    \node[urbox,text width=1.95cm] (request) at (640,715)
      {{\bfseries User request}\\[1.5pt]{\itshape\fontsize{6.6}{7.6}\selectfont ``Introduce the car in front of me''}};

    \fill[white,rounded corners=8pt] (10,824) rectangle (1470,878);
    \draw[CleS2bd,line width=1.2pt,rounded corners=8pt] (10,824) rectangle (1470,878);
    \node[font=\footnotesize,text=LocInk] at (740,851) {One shared
      \texttt{\textbf{TraceRecord}} identity across \textcolor{LocS2}{\textbf{System~2}},
      \textcolor{LocS1}{\textbf{System~1}}, \textcolor{LocS0}{\textbf{System~0}}, and
      \textcolor{CleUR}{\textbf{user}} events};

    \begin{pgfonlayer}{cle-wires}
      \draw[thread] (task.south) -- (navuser.north);
      \draw[thread] (navuser.east) -- (greeting.west);
      \draw[thread] (resolve.east) -- (knowledge.west);
      \draw[thread] (resolve.south) -- (850,330) -- (970,330) -- (navveh.north);
      \draw[thread] (knowledge.south) -- (vehintro.north);
      \draw[thread] (navveh.east) -- (vehintro.west);
      \draw[thread] (vehintro.east) -- (presmot.west);
      \draw[s0wire] (navuser.south) -- (loc1.north);
      \draw[s0wire] (greeting.south) -- (wbc.north);
      \draw[s0wire] (navveh.south) -- (loc2.north);
      \draw[s0wire] (presmot.south) -- (motgen.north);
      \draw[thread] (greeting.north) -- (510,330) -- (585,330) -- (585,0 |- dialogue.south);
      \draw[userwire] (640,0 |- dialogue.south) -- (640,0 |- request.north);
      \draw[userwire] (request.east) -- (805,715) -- (805,330) -- (850,330) -- (850,0 |- resolve.south);
    \end{pgfonlayer}
  \end{tikzpicture}}
  
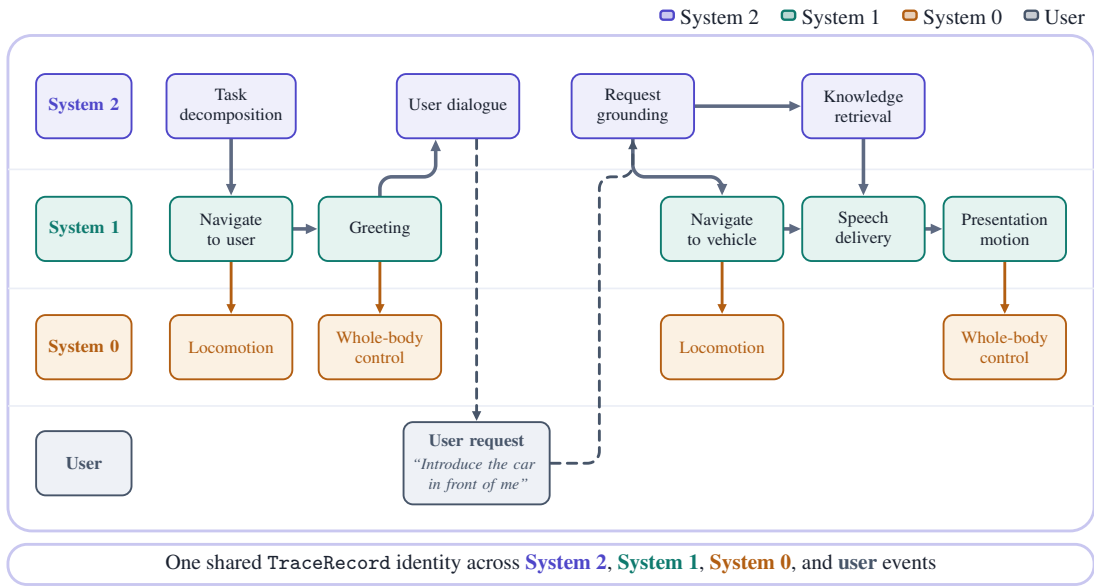
\captionof{figure}{Cross-layer episode execution for the interactive vehicle-guide case.
The swim lanes separate System~2 reasoning and dialogue, System~1 skills,
System~0 support, and user events. Arrows mark the main execution flow,
System~0 support links, and user-response flow. All events are recorded under
one TraceRecord identity, enabling end-to-end evaluation and failure
attribution.}
  \label{fig:full-system-timeline}
\end{minipage}
\end{center}

\paragraph{Full-system evaluator.}
For full-system validation, DeepInsight provides a trace-based evaluation
mechanism that maps one composed execution trace to episode-, subgoal-, and
subsystem-level judgments. In this representative task, the episode-level
judgment reports end-to-end task success. The subgoal-level judgment evaluates
whether the semantic, interactive, navigational, and physical criteria of the
task are satisfied along the episode, from recognizing the user's assistance
intent to reaching the referenced object and completing the presentation. The
subsystem-level judgment reports the supporting subsystem signals,
including System~2 tool and dialogue decisions, System~1 outcomes for
navigation and presentation delivery, and System~0 locomotion
and whole-body-control stability.
Because these judgments are derived from the same trace identity rather than
from separate evaluation scripts, the episode score, subgoal scores, and
subsystem diagnostics remain grounded in the same episode events. In the
DeepInsight abstraction, the task defines the episode contract, the resource
binds the agent to executable skills and control, and the result turns the
composed trace into hierarchical judgments.

\paragraph{End-to-end decomposition.}
A central output of the full-system evaluator is the decomposition of the gap
between subgoal-level completion and end-to-end task success.
Table~\ref{tab:full-system-component-gap} summarizes the results over
96 episodes across three layout groups: 58 episodes completed
the task end-to-end, while 38 failed to satisfy the full success
criterion. Each
row corresponds to an evaluation criterion associated with a task step,
subsystem signal, or cross-step state dependency. Criterion-level rates are
marginal diagnostics over all 96 episodes: failures can overlap across rows, and
the final row requires all required criteria to hold within the same episode.
The results expose a clear composition gap: high local completion rates do not
necessarily yield a successful system-level outcome. At the same time, the
criterion-level view makes weak points observable, showing where performance is
lost before the episode-level failure is emitted.

\begin{center}
\begin{minipage}{\linewidth}
  \centering
  \captionof{table}{Subgoal-level report for the interactive vehicle-guide task over 96
  episodes. Each criterion row is a marginal diagnostic evaluated from the shared
  episode trace and reports one task criterion, its supporting evidence, the
  associated system locus, and the criterion-level success rate. The final row
  reports end-to-end success when all required criteria are satisfied within the
  same episode.}
  \label{tab:full-system-component-gap}
  \footnotesize
  \setlength{\tabcolsep}{4pt}
  \renewcommand{\arraystretch}{0.93}
  \begin{tabularx}{\linewidth}{>{\raggedright\arraybackslash}X
                              >{\raggedright\arraybackslash}p{0.30\linewidth}
                              >{\raggedright\arraybackslash}p{0.17\linewidth}
                              >{\centering\arraybackslash}p{0.11\linewidth}}
    \toprule
    Evaluation criterion & Evidence & Locus & Success (\%) \\
    \midrule
    Task intent recognized & instruction is parsed as a user-first interactive guide task & System~2 & 94.8 \\
    Tool sequence valid & required calls appear in order with legal arguments & System~2 / tool contracts & 93.8 \\
    User target selected & vision-search navigation returns the intended user candidate & System~1 navigation & 82.3 \\
    User reached & final distance and yaw satisfy the social-interaction condition & System~1 + System~0 & 77.1 \\
    Greeting completed & wave action finishes and the controller reports stable posture & System~1 + System~0 & 83.3 \\
    User request consumed & user-agent event is joined into the next semantic analysis & System~2 dialogue state & 91.7 \\
    Vehicle-guide request grounded & user request is grounded as a navigation-and-presentation subtask for the referenced vehicle & System~2 & 90.6 \\
    Vehicle target selected & navigation selects the correct car from candidate objects & System~1 navigation & 74.0 \\
    Vehicle reached & final pose satisfies distance, view, and presentation requirements & System~1 + System~0 & 70.8 \\
    Post-arrival state handed off & navigation result provides the vehicle state for subsequent reasoning & System~1 / System~2 handoff & 85.4 \\
    Presentation content valid & request intent is recognized and vehicle-specific presentation content is generated & System~2 & 93.8 \\
    Presentation motion aligned & generated motion has sync points for the explanation & System~2 / System~1 handoff & 82.3 \\
    Presentation executed & speech and whole-body presentation complete without losing the valid pose & System~0 whole-body & 91.7 \\
    Collision-free execution & trace contains no physical contact violation with scene objects & System~0 safety & 82.3 \\
    \midrule
    End-to-end episode success & all required criteria are satisfied within the same episode & Full system & 60.4 \\
    \bottomrule
  \end{tabularx}
\end{minipage}
\end{center}

\paragraph{Failure attribution.}
Beyond task success, full-system validation must identify where performance is
lost. For each non-success episode, a judge agent assigns the episode a primary
failure label from violated criteria, execution traces, tool statuses, and
control events. This attribution makes both subsystem failures and
system-boundary handoff failures observable: a navigation event may succeed
locally while violating the downstream task precondition, a selected route may be
difficult for locomotion to execute, and generated presentation motion may
exceed whole-body-control feasibility constraints. We manually inspect a subset
of the attributions for quality control. Figure~\ref{fig:full-system-failure-locus}
aggregates the 38 non-success episodes by failure locus, and
Table~\ref{tab:full-system-failures} expands the corresponding primary labels.

\newcommand{\locsw}[1]{\raisebox{0.04em}{\tikz{\fill[#1, rounded corners=0.4pt]
  (0,0) rectangle (0.62em,0.62em);}}}
\begin{center}
  \begin{minipage}[c]{0.40\linewidth}
    \centering
    \begin{tikzpicture}[font=\scriptsize]
      \def\r{1.55}
      \def\ir{0.92}
      \path[fill=LocS1, draw=white, line width=1.0pt]
        (88.9:\r) arc[start angle=88.9,end angle=-60.48,radius=\r] --
        (-60.48:\ir) arc[start angle=-60.48,end angle=88.9,radius=\ir] -- cycle;
      \path[fill=LocHandoff, draw=white, line width=1.0pt]
        (-62.68:\r) arc[start angle=-62.68,end angle=-164.69,radius=\r] --
        (-164.69:\ir) arc[start angle=-164.69,end angle=-62.68,radius=\ir] -- cycle;
      \path[fill=LocS2, draw=white, line width=1.0pt]
        (-166.89:\r) arc[start angle=-166.89,end angle=-231.01,radius=\r] --
        (-231.01:\ir) arc[start angle=-231.01,end angle=-166.89,radius=\ir] -- cycle;
      \path[fill=LocS0, draw=white, line width=1.0pt]
        (-233.21:\r) arc[start angle=-233.21,end angle=-268.90,radius=\r] --
        (-268.90:\ir) arc[start angle=-268.90,end angle=-233.21,radius=\ir] -- cycle;
      \node[align=center, text=LocInk] at (0,0)
        {{\fontsize{20}{20}\selectfont\bfseries 38}\\[1pt]
         {\color{LocInkSoft}\scriptsize non-success}};
    \end{tikzpicture}
  \end{minipage}\hfill
  \begin{minipage}[c]{0.56\linewidth}
    \small
    \setlength{\tabcolsep}{6pt}
    \renewcommand{\arraystretch}{1.35}
    \begin{tabularx}{\linewidth}{@{}cX r@{\hspace{0.5em}}r@{}}
      \multicolumn{4}{@{}l}{\footnotesize\bfseries\color{LocInkSoft}\MakeUppercase{Failure locus}}\\[2pt]
      \cmidrule[0.4pt]{1-4}\noalign{\vskip2pt}
      \locsw{LocS1}      & System~1  & \texttt{16} & \texttt{42.1\%}\\
      \locsw{LocHandoff} & Boundary handoff & \texttt{11} & \texttt{28.9\%}\\
      \locsw{LocS2}      & System~2  & \texttt{\phantom{0}7} & \texttt{18.4\%}\\
      \locsw{LocS0}      & System~0  & \texttt{\phantom{0}4} & \texttt{10.5\%}\\
    \end{tabularx}
  \end{minipage}
  
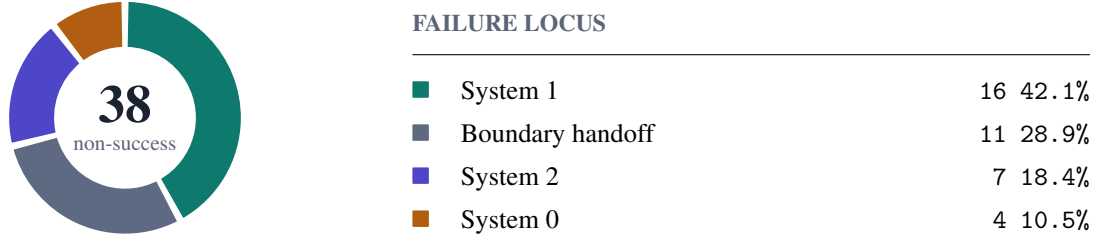
\captionof{figure}{Primary failure-locus distribution over the 38 non-success
  episodes, ordered by descending share. Each episode is assigned one primary
  failure locus by the judge agent. Boundary handoff denotes failures at system
  boundaries during composed execution, including mismatched state,
  preconditions, timing, success semantics, route executability, or controller
  feasibility across adjacent systems.}
  \label{fig:full-system-failure-locus}
\end{center}

\begin{center}
\begin{minipage}{\linewidth}
  \centering
  \captionof{table}{Detailed failure labels for the non-success episodes in
  Figure~\ref{fig:full-system-failure-locus}.}
  \label{tab:full-system-failures}
  \scriptsize
  \setlength{\tabcolsep}{3.2pt}
  \renewcommand{\arraystretch}{1.02}
  \begin{tabularx}{\linewidth}{>{\raggedright\arraybackslash}p{0.22\linewidth}
                              >{\raggedright\arraybackslash}p{0.12\linewidth}
                              >{\raggedright\arraybackslash}p{0.15\linewidth}
                              >{\centering\arraybackslash}p{0.08\linewidth}
                              >{\centering\arraybackslash}p{0.09\linewidth}
                              >{\raggedright\arraybackslash}X}
    \toprule
    Primary failure label & Failure class & Locus & Count & Share (\%) & Diagnostic trace evidence \\
    \midrule
    User target-search failure & module & System~1 navigation & 5 & 13.2 & vision-search navigation misses the user, selects a false positive, or times out \\
    Vehicle target-search failure & module & System~1 navigation & 5 & 13.2 & referenced vehicle is not selected from candidate objects after the request is grounded \\
    Vehicle navigation execution failure & module & System~1 navigation & 6 & 15.8 & target vehicle is correct, but path planning, progress, or arrival timeout fails \\
    Navigation-to-task boundary mismatch & boundary & System~1 / System~2 & 3 & 7.9 & navigation reports local success, but pose, visibility, or post-arrival state does not satisfy the next task precondition \\
    Navigation-to-locomotion executability mismatch & boundary & System~1 / System~0 & 3 & 7.9 & navigation selects the correct route or waypoint sequence, but the path is difficult for locomotion to execute under physical constraints \\
    Task scheduling / grounding error & module & System~2 & 5 & 13.2 & assistance intent or vehicle-guide request is not grounded into the correct navigation-and-presentation subtask \\
    Control execution / safety error & control & System~0 & 4 & 10.5 & collision with scene objects or whole-body instability invalidates the rollout \\
    Dialogue state-handling error & module & System~2 & 2 & 5.3 & user reply is not joined into the next semantic analysis \\
    Motion-to-WBC executability mismatch & boundary & System~1 / System~0 & 3 & 7.9 & generated presentation motion is difficult for whole-body control to execute under balance, range, or timing constraints \\
    Success-contract boundary mismatch & boundary & System~1 / System~2 & 2 & 5.3 & System~1 tool reports local success, but the shared-trace criterion needed by System~2 remains unsatisfied \\
    \bottomrule
  \end{tabularx}
\end{minipage}
\end{center}

The case study is not intended to introduce a new benchmark; it isolates the
part of Physical AI evaluation that component benchmarks cannot measure. The
results show that system-level task performance is not reducible to isolated
module-level quality: the largest failure locus lies in System~1 execution
(16/38), driven mainly by user/vehicle search and vehicle-navigation failures,
while system-boundary handoff forms the second-largest locus (11/38), followed by
System~2 reasoning failures (7/38).
These boundary failures expose latent gaps left by independently designed
subsystems: assumptions about state, preconditions, timing, and executability
that are not visible when each module is evaluated in isolation, but that
strongly affect the final system outcome. In this sense, full-system validation
turns architecture into an evaluation target and reveals the integration gaps
that determine the practical performance ceiling of the embodied system.

\section{Conclusion}
\label{sec:conclusion}

DeepInsight brings the embodied humanoid stack---language reasoning through
whole-body control---onto a single runtime, in place of the patchwork of
incompatible harnesses each layer would otherwise demand. In production it
onboards new benchmarks largely through configuration, reproduces matched
System~2 references where mature peer frameworks exist, and carries the same
runtime abstractions into release-oriented embodied evaluation. Where peer
frameworks exist it stays competitive on a single node---and, unlike them,
scales across multiple nodes without re-tuning. Its deeper return is diagnostic:
because every layer writes into one shared trace, a regression that begins in
one layer and surfaces in another stays localizable on that trace---the
cross-layer diagnosis that a federation of separate harnesses, however well
coordinated, cannot reconstruct.

The work ahead is to broaden the embodied side of the stack. First, System~1
and System~0 need wider task-family coverage: more navigation, manipulation,
motion-generation, locomotion, safety, and whole-body-control evaluations should
enter through the same task and result interfaces, so that the production
surface is not limited to the representative case studies reported here.
Second, every rollout reported here is simulated, and the sim-to-real gap
remains the decisive uncertainty for policies bound for hardware. A natural
next step is therefore to put real-world robots behind the same resource-handle
protocol as simulators, allowing simulated and physical rollouts to share one
trace identity. That would turn the sim-to-real gap from an external deployment
risk into something DeepInsight can measure, compare, and diagnose directly.

\bibliographystyle{unsrtnat}
\IfFileExists{refs.bib}{\bibliography{refs}}{}

\clearpage
\appendix

\section{System~1 Modality and Extensibility Details}
\label{app:system1-details}

\begin{table}[!htbp]
  \centering
  \caption{Audio-conditioned motion-generation preference evaluation over
  20 clips. Ten blind raters compare Model~A and Model~B on each dimension;
  model identity and presentation order are randomized and hidden from raters.
  Dimension rows report categorical judgment counts; the aggregate row reports
  pooled percentages across all dimensions.}
  \label{tab:exp-s1-motion}
  \small
  \setlength{\tabcolsep}{5pt}
  \renewcommand{\arraystretch}{1.05}
  \begin{tabular}{lrrrrr}
    \toprule
    Dimension & A better & B better & Both good & Both poor & Total \\
    \midrule
    Scene / content match              & 96 & 46 & 41 & 17 & 200 \\
    Motion completeness and smoothness & 62 & 51 & 74 & 13 & 200 \\
    Perceived safety                   & 42 & 35 & 112 & 11 & 200 \\
    Expressive style                   & 90 & 48 & 28 & 34 & 200 \\
    \midrule
    Aggregate (\%)                     & 36.3 & 22.5 & 31.9 & 9.4 & 100.0 \\
    \bottomrule
  \end{tabular}
\end{table}

\begin{table}[!htbp]
  \centering
  \caption{System~1 evaluation extensibility by architectural boundary.
  Benchmark-specific requirements are localized to boundary bindings, while the
  unified simulation, execution, model-interface, and reporting abstractions are
  reused.}
  \label{tab:exp-s1-extensibility}
  \small
  \setlength{\tabcolsep}{5pt}
  \renewcommand{\arraystretch}{1.08}
  \begin{tabularx}{\textwidth}{>{\raggedright\arraybackslash}p{0.20\textwidth}>{\raggedright\arraybackslash}X>{\raggedright\arraybackslash}X}
    \toprule
    Boundary & Reused infrastructure & Benchmark-specific binding \\
    \midrule
    Scene / data binding &
    scene abstraction, coordinate conventions, asset representation &
    scene content, raw data format, maps, semantic labels, asset metadata \\
    Task adapter &
    episode driver, interaction loop, termination handling &
    episode specification, observations, actions, goal specification \\
    Execution binding &
    rollout abstraction, resource model, simulator integration &
    benchmark-specific execution parameters and embodiment settings \\
    Model binding &
    model endpoint abstraction, inference contract, adapter interface &
    model-specific observation/action adapter \\
    Evaluation / reporting &
    trace schema, aggregation protocol, report abstraction &
    benchmark metrics and evaluation reducers \\
    \bottomrule
  \end{tabularx}
\end{table}

\section{System~0 Screening and Diagnostic Details}
\label{app:system0-details}

\begin{table}[!htbp]
\centering
\vspace{-0.5em}
\footnotesize
\setlength{\tabcolsep}{3pt}
\renewcommand{\arraystretch}{0.96}
\caption{System~0 model-pool leaderboard excerpt for the policy-set screen.
Rows are anonymized checkpoint entries. Columns keep only report-visible
model-pool fields, aggregate metrics used for ranking, training settings, and
brief checkpoint notes; the full SR--MPJPE distribution is visualized in
\figref{fig:system0-policy-plane}.}
\label{tab:system0-policy-set}
\begin{tabularx}{\textwidth}{>{\raggedright\arraybackslash}p{0.12\textwidth}>{\raggedright\arraybackslash}p{0.13\textwidth}>{\centering\arraybackslash}p{0.12\textwidth}>{\raggedleft\arraybackslash}p{0.08\textwidth}>{\raggedleft\arraybackslash}p{0.10\textwidth}>{\raggedright\arraybackslash}p{0.15\textwidth}>{\raggedright\arraybackslash}X}
\toprule
Model ID & Model name & Release state & SR (\%) & MPJPE (cm) & Training Config & Notes \\
\midrule
\texttt{s0\_001} & \textsc{WBC-RC-01} & Blocked & \textbf{97.90} & \textbf{1.915} & Data v3 / reward v4 / sampler v2 & Reward update \\
\texttt{s0\_002} & \textsc{WBC-RC-02} & Not advanced & 97.53 & 2.178 & Data v3 / reward v4 / distill v1 & Distillation \\
\texttt{s0\_003} & \textsc{WBC-RC-03} & Not advanced & 97.44 & 2.198 & Data v4 / reward v3 / sampler v1 & Dataset refresh \\
\bottomrule
\end{tabularx}
\renewcommand{\arraystretch}{1.0}
\vspace{-0.4em}
\end{table}

\vspace{0.15em}
\begin{center}
\begin{minipage}{\textwidth}
\centering
\small
\setlength{\fboxsep}{2pt}
\begin{minipage}[t]{0.49\linewidth}
  \centering
  \IfFileExists{figures/system0_screen_video1_t12.png}{%
    \includegraphics[width=\linewidth]{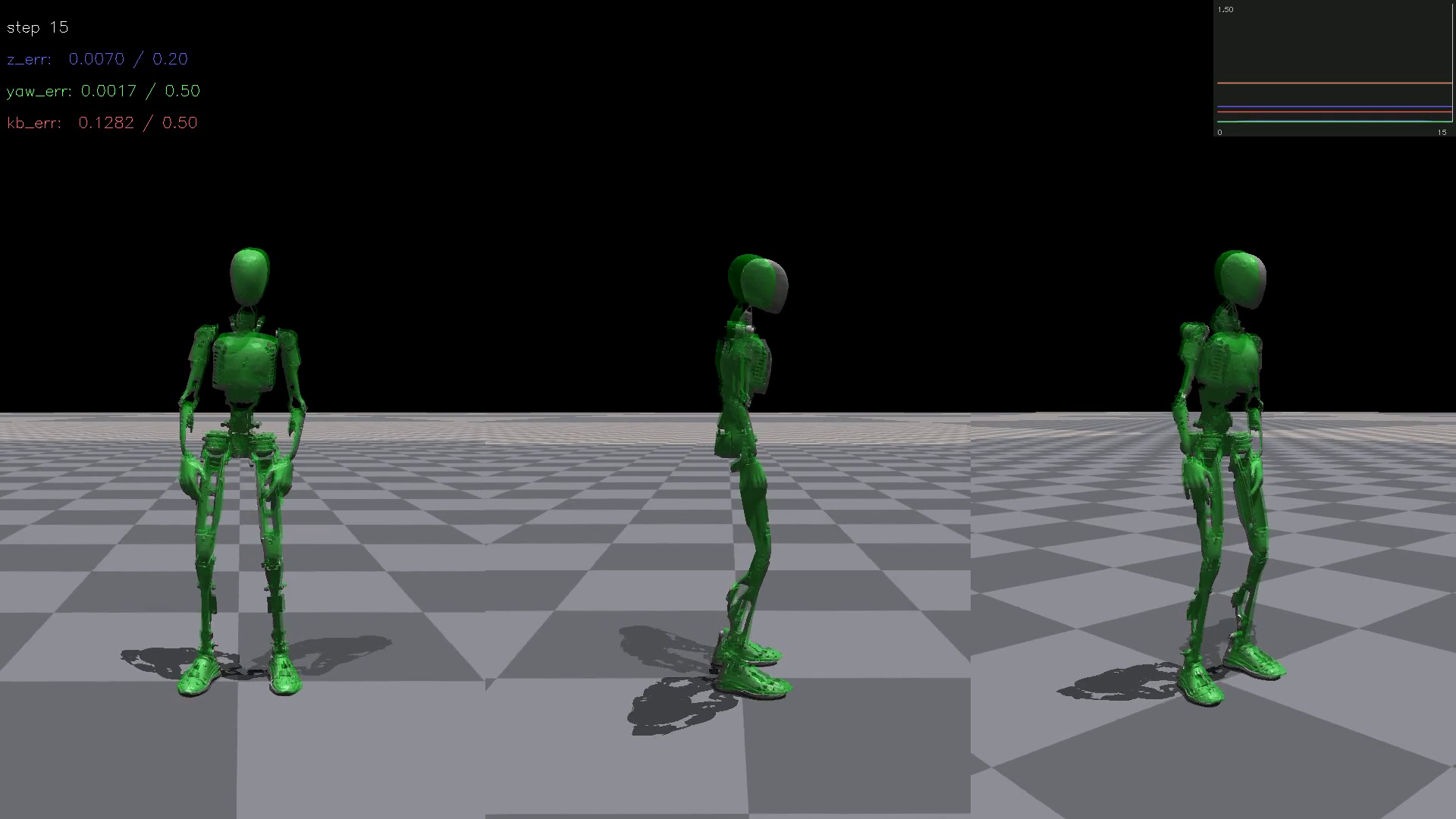}%
  }{%
    \fbox{\begin{minipage}[c][0.26\linewidth][c]{0.94\linewidth}\centering\textbf{Placeholder}\\Screening video still\end{minipage}}%
  }
  \vspace{0.1em}
  {\scriptsize (a) High-dynamic clip A, early rollout}
\end{minipage}\hfill
\begin{minipage}[t]{0.49\linewidth}
  \centering
  \IfFileExists{figures/system0_screen_video1_t48.png}{%
    \includegraphics[width=\linewidth]{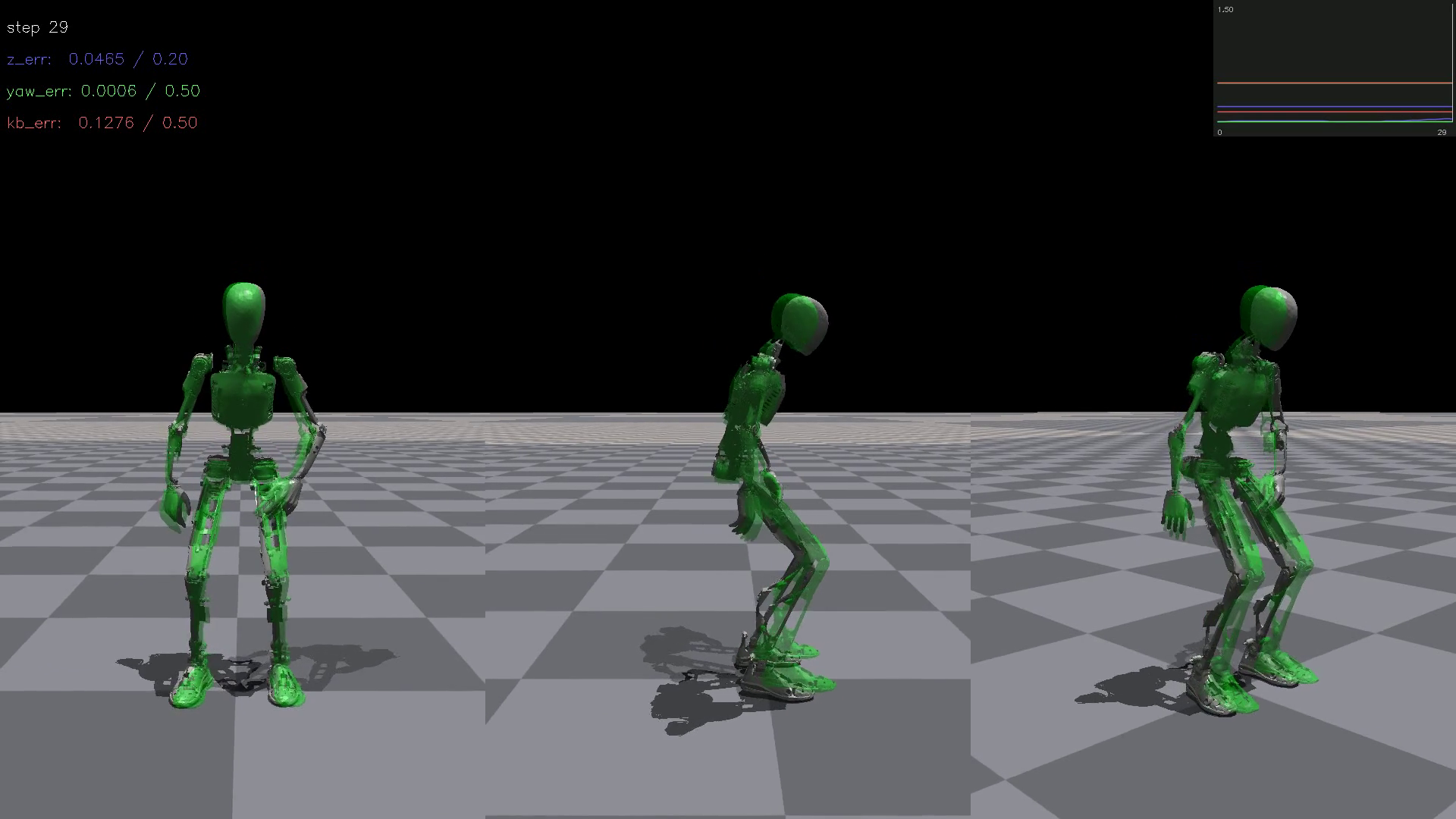}%
  }{%
    \fbox{\begin{minipage}[c][0.26\linewidth][c]{0.94\linewidth}\centering\textbf{Placeholder}\\Screening video still\end{minipage}}%
  }
  \vspace{0.1em}
  {\scriptsize (b) High-dynamic clip A, later rollout}
\end{minipage}

\vspace{0.35em}
\begin{minipage}[t]{0.49\linewidth}
  \centering
  \IfFileExists{figures/system0_screen_video2_t14.png}{%
    \includegraphics[width=\linewidth]{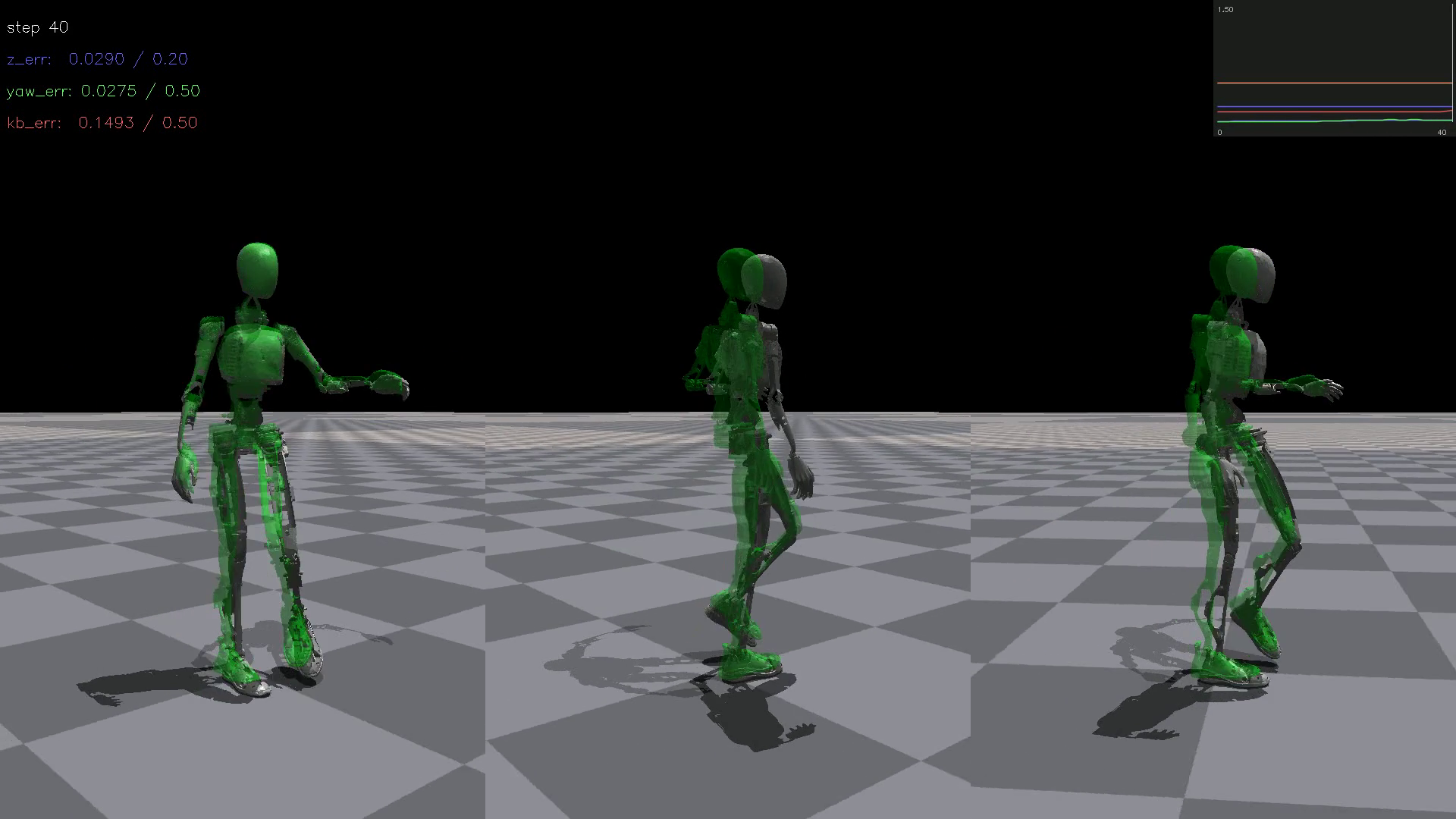}%
  }{%
    \fbox{\begin{minipage}[c][0.26\linewidth][c]{0.94\linewidth}\centering\textbf{Placeholder}\\Screening video still\end{minipage}}%
  }
  \vspace{0.1em}
  {\scriptsize (c) High-dynamic clip B, early rollout}
\end{minipage}\hfill
\begin{minipage}[t]{0.49\linewidth}
  \centering
  \IfFileExists{figures/system0_screen_video2_t52.png}{%
    \includegraphics[width=\linewidth]{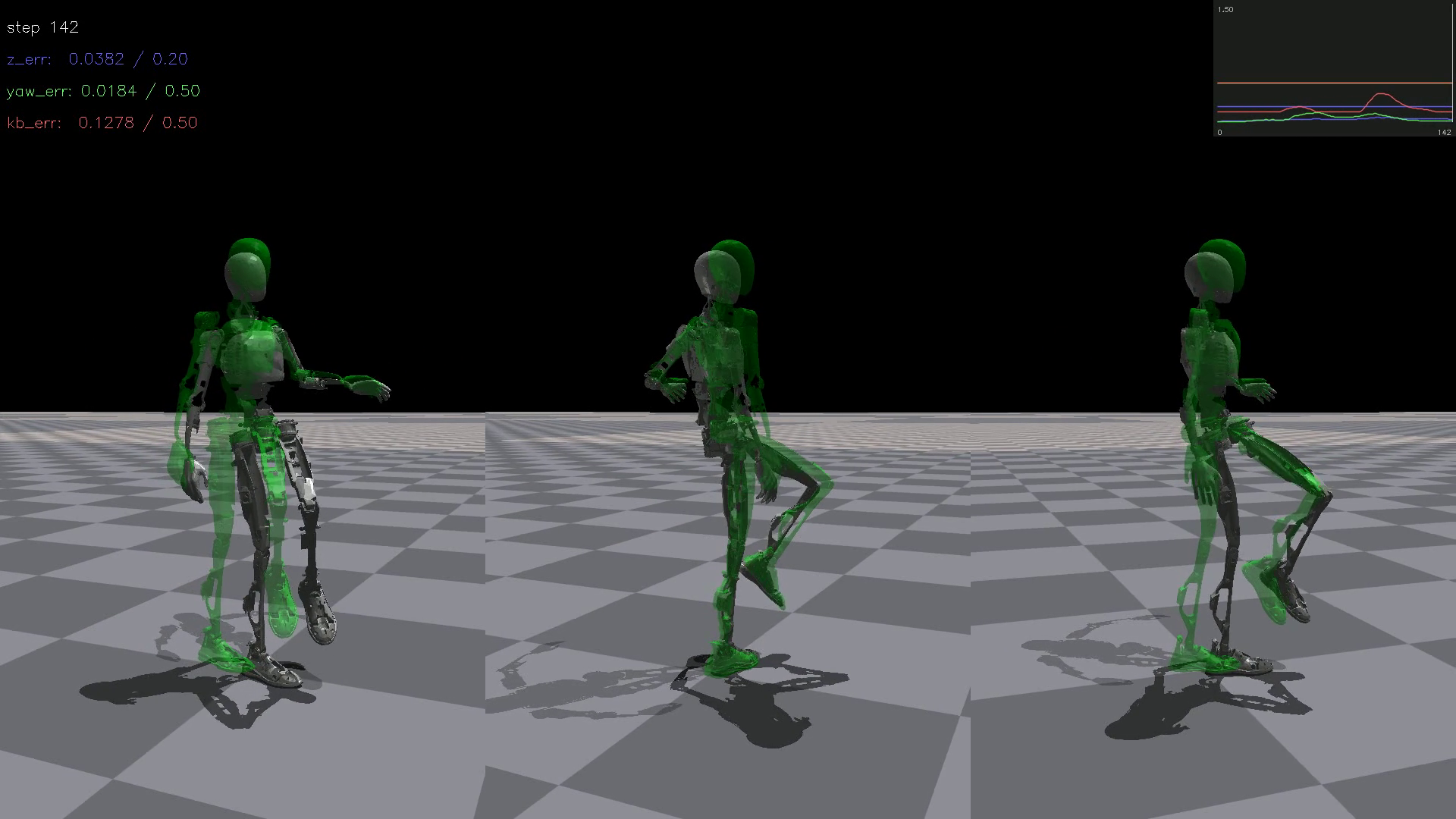}%
  }{%
    \fbox{\begin{minipage}[c][0.26\linewidth][c]{0.94\linewidth}\centering\textbf{Placeholder}\\Screening video still\end{minipage}}%
  }
  \vspace{0.1em}
  {\scriptsize (d) High-dynamic clip B, later rollout}
\end{minipage}
\vspace{-0.35em}
\captionsetup{skip=4pt}
\captionof{figure}{Representative stills from high-dynamic policy-set evaluation videos attached to
the model-pool run. Green overlays show the replay/reference motion and
gray--white overlays show the evaluated checkpoint output across synchronized
views. DeepInsight batches the broader clip set across candidate policies,
computes SR and MPJPE from logged rollout metrics, and retains the videos as
auditable evidence for selected model--clip cases.}
\label{fig:system0-screening-frames}
\end{minipage}
\end{center}
\vspace{-0.35em}

\begin{table}[!htbp]
\centering
\vspace{-0.4em}
\footnotesize
\setlength{\tabcolsep}{3.5pt}
\renewcommand{\arraystretch}{0.98}
\caption{Compact selected-policy diagnostic metric schema. The table
groups representative statistics by the behavior family they
support; pass/fail decisions are reported separately in
\tabref{tab:system0-mujoco-diagnostics}.}
\label{tab:system0-diagnostic-schema}
\begin{tabularx}{\textwidth}{>{\raggedright\arraybackslash}p{0.22\textwidth}>{\raggedright\arraybackslash}X>{\raggedright\arraybackslash}p{0.24\textwidth}}
\toprule
Metric family & Representative statistics & Diagnostic purpose \\
\midrule
Command tracking & Linear-velocity residuals; yaw-rate tracking residuals & Verify that commanded motion is followed \\
Gait and style & Step period; step length; left--right asymmetry & Detect unstable or asymmetric locomotion patterns \\
Swing-foot clearance & Mean foot clearance; bilateral clearance imbalance & Detect foot-dragging and clearance imbalance \\
Joint dynamics & Mean hip-pitch angular velocity; bilateral hip-dynamics imbalance & Detect localized joint-dynamics abnormalities \\
Contact and posture & Touchdown foot-pitch angle; torso pitch angle & Detect contact-attitude and body-posture risk \\
Upper-body kinematics & Elbow- and shoulder-pitch range of motion & Detect frozen or excessive upper-body motion \\
Style scores & Forward, backward, turn, and rotate imitation scores & Provide imitation-style context across motion modes \\
\bottomrule
\end{tabularx}
\renewcommand{\arraystretch}{1.0}
\vspace{-0.4em}
\end{table}

\FloatBarrier
\begin{center}
\begin{minipage}{\textwidth}
\centering
\small
\setlength{\fboxsep}{2pt}
\begin{minipage}[t]{0.235\linewidth}
  \centering
  \IfFileExists{figures/system0_diag_hip_dynamics.png}{%
    \includegraphics[width=\linewidth]{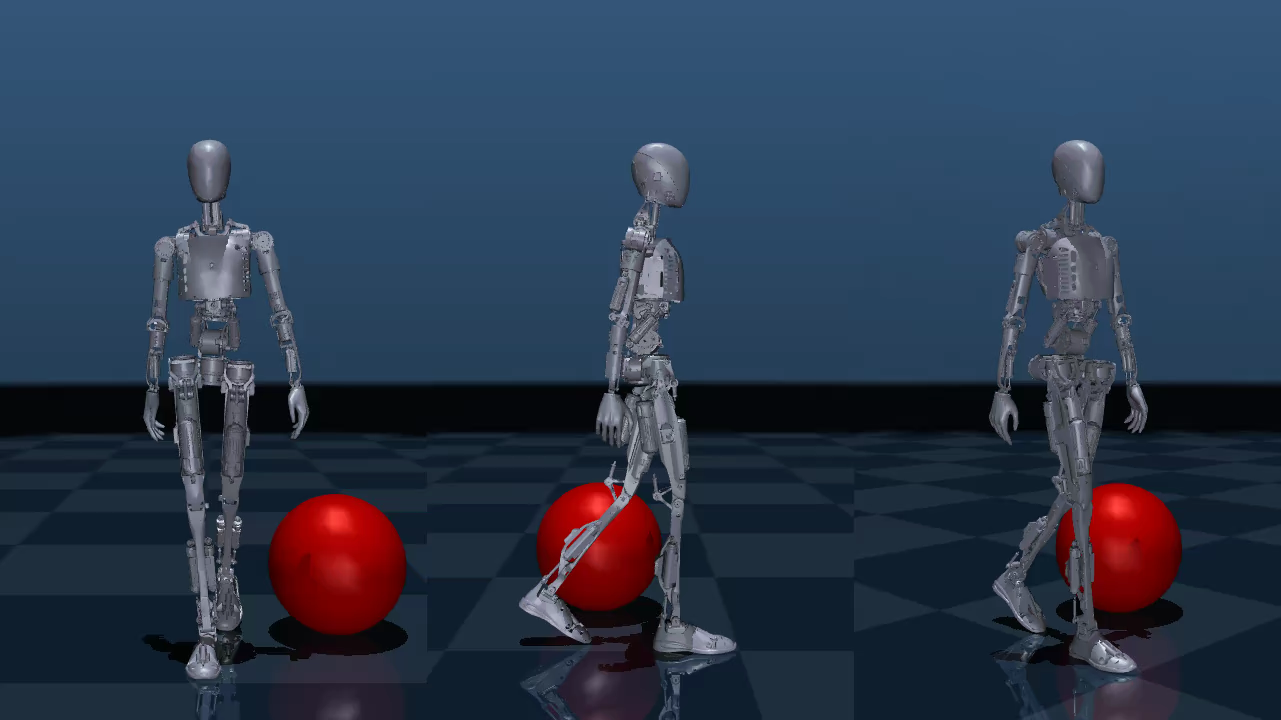}%
  }{%
    \fbox{\begin{minipage}[c][0.56\linewidth][c]{0.94\linewidth}\centering\textbf{Placeholder}\\Hip dynamics\end{minipage}}%
  }
  \vspace{0.1em}
  {\scriptsize (a) $t=5.00$~s}
\end{minipage}\hfill
\begin{minipage}[t]{0.235\linewidth}
  \centering
  \IfFileExists{figures/system0_diag_command_tracking.png}{%
    \includegraphics[width=\linewidth]{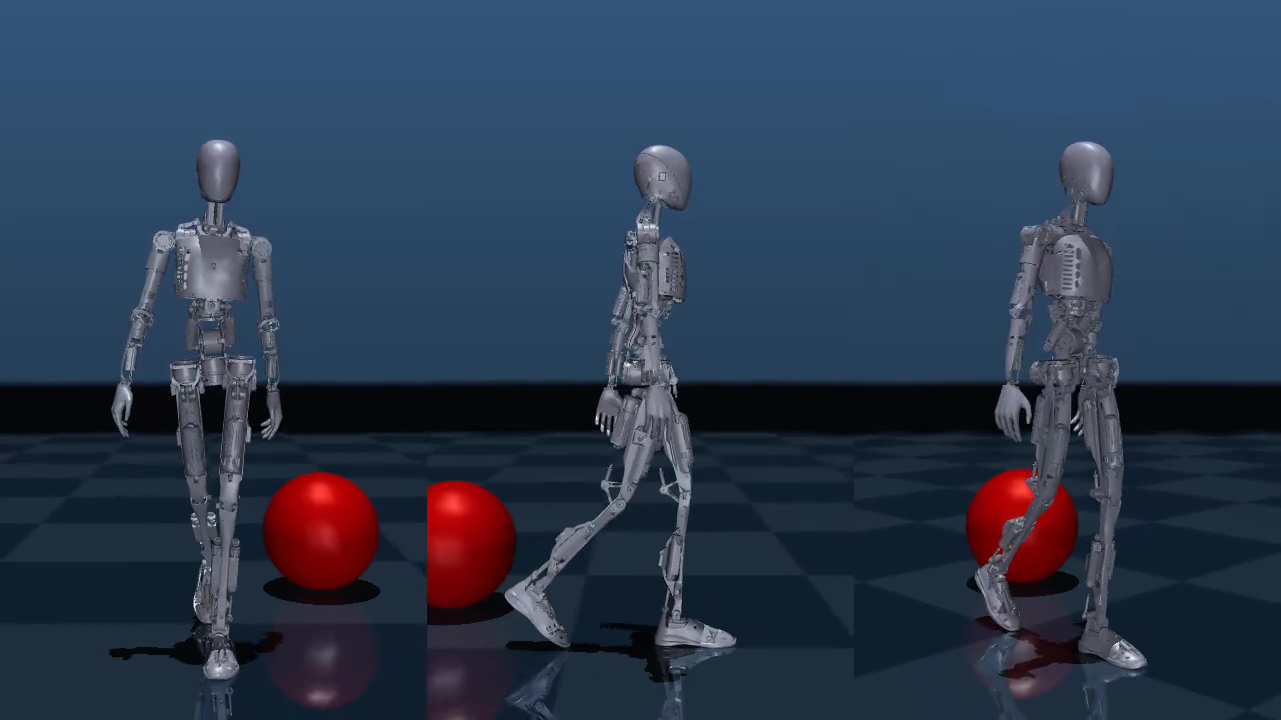}%
  }{%
    \fbox{\begin{minipage}[c][0.56\linewidth][c]{0.94\linewidth}\centering\textbf{Placeholder}\\Tracking\end{minipage}}%
  }
  \vspace{0.1em}
  {\scriptsize (b) $t=5.75$~s}
\end{minipage}\hfill
\begin{minipage}[t]{0.235\linewidth}
  \centering
  \IfFileExists{figures/system0_diag_contact_posture.png}{%
    \includegraphics[width=\linewidth]{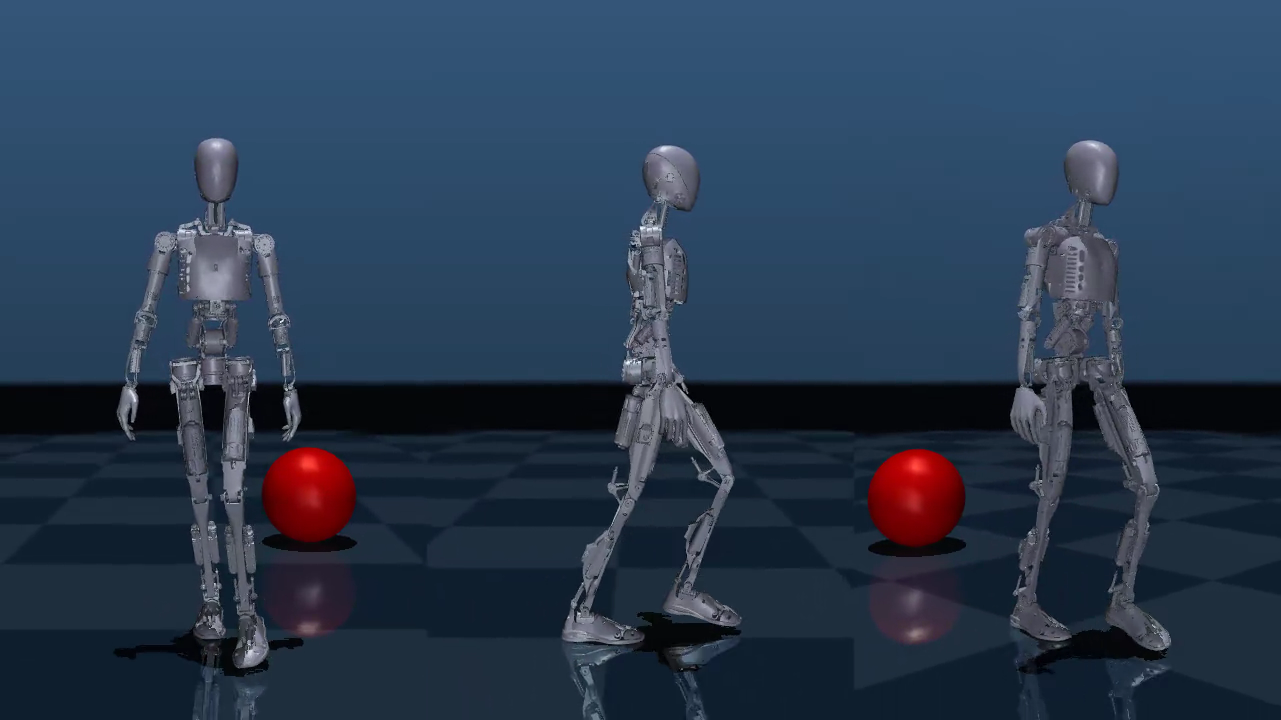}%
  }{%
    \fbox{\begin{minipage}[c][0.56\linewidth][c]{0.94\linewidth}\centering\textbf{Placeholder}\\Contact posture\end{minipage}}%
  }
  \vspace{0.1em}
  {\scriptsize (c) $t=7.00$~s}
\end{minipage}\hfill
\begin{minipage}[t]{0.235\linewidth}
  \centering
  \IfFileExists{figures/system0_diag_upper_body.png}{%
    \includegraphics[width=\linewidth]{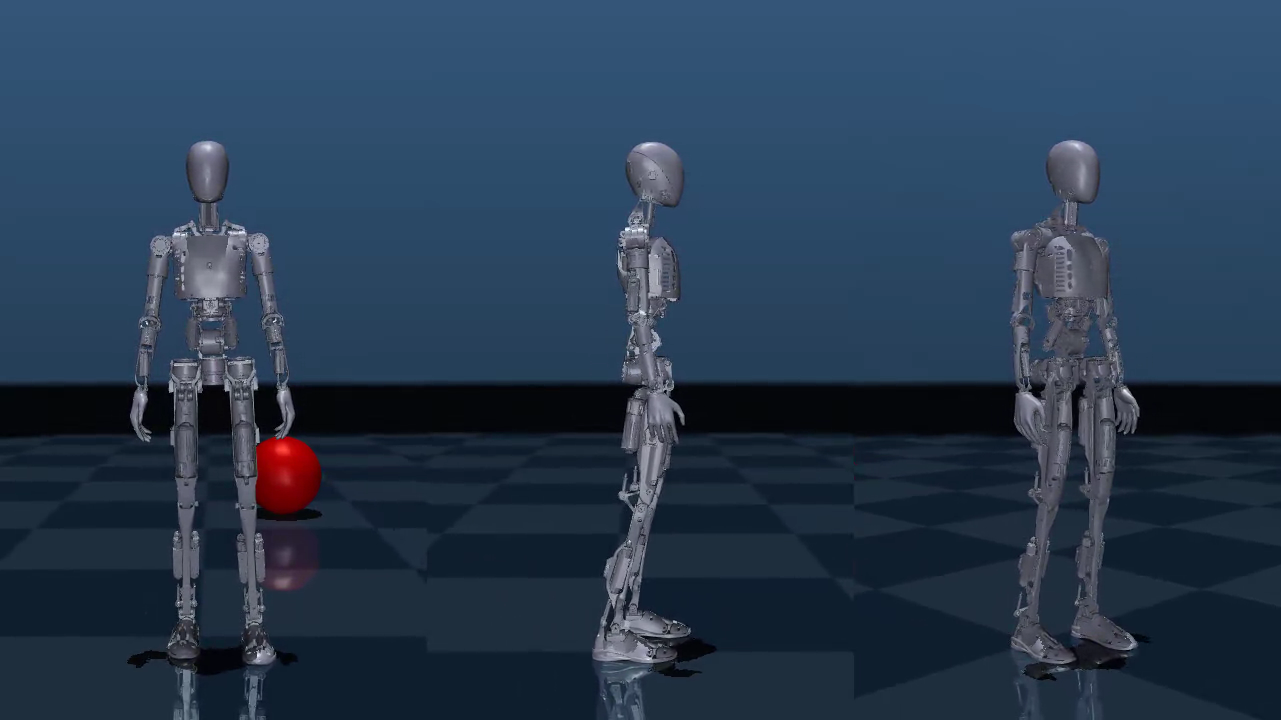}%
  }{%
    \fbox{\begin{minipage}[c][0.56\linewidth][c]{0.94\linewidth}\centering\textbf{Placeholder}\\Upper body\end{minipage}}%
  }
  \vspace{0.1em}
  {\scriptsize (d) $t=9.50$~s}
\end{minipage}
\vspace{-0.4em}
\captionsetup{skip=4pt}
\captionof{figure}{Representative time-ordered stills from the selected-policy diagnostic video. The
video provides visual context for the diagnostic checks;
\tabref{tab:system0-mujoco-diagnostics} reports the corresponding pass/fail
outcomes used by the release decision.}
\label{fig:system0-mujoco-frames}
\end{minipage}
\end{center}

\end{document}